\documentclass[11pt]{article}

\usepackage[in]{fullpage}
\usepackage[utf8]{inputenc} 
\usepackage[T1]{fontenc}    
\usepackage{hyperref}       
\usepackage{url}            
\usepackage{booktabs}       
\usepackage{amsfonts}       
\usepackage{nicefrac}       
\usepackage{microtype}      

\usepackage{graphicx,verbatim}
\usepackage[rflt]{floatflt}
\usepackage{epsfig,times,subfigure,graphicx}
\usepackage{amsmath,amssymb,amsopn,algorithm,algorithmic,theorem,float,bbm,bm,enumerate,color,multirow}
\usepackage{rotating}
\usepackage{array}
\usepackage{hyperref}
\usepackage{url}
\usepackage{mathtools}
\usepackage{chngcntr}
\usepackage{nameref}
\usepackage{zref-xr}
\zxrsetup{toltxlabel}

\newcommand{\beq}{\begin{equation}}
\newcommand{\eeq}{\end{equation}}

\newcommand\B{\mathbb{B}}

\newcommand\s{\mathbb{S}}
\newcommand\R{\mathbb{R}}

\renewcommand\P{\mathbb{P}}

\newcommand{\g}{\mathbf{g}}

\renewcommand{\a}{\mathbf{a}}

\renewcommand{\u}{\mathbf{u}}
\renewcommand{\v}{\mathbf{v}}

\newcommand{\w}{\mathbf{w}}
\newcommand{\x}{\mathbf{x}}
\newcommand{\y}{\mathbf{y}}
\newcommand{\z}{\mathbf{z}}

\renewcommand{\t}{\mathbf{t}}

\newcommand{\cH}{{\cal H}}

\newcommand{\cN}{{\cal N}}
\newcommand{\cP}{{\cal P}}
\newcommand{\cQ}{{\cal Q}}

\newcommand{\cT}{{\cal T}}

\newcommand{\cB}{{\cal B}}
\newcommand{\cD}{{\cal D}}
\newcommand{\cA}{{\cal A}}

\newcommand{\bI}{\mathbf{I}}

\newcommand{\bX}{\mathbf{X}}

\newcommand{\vertiii}[1]{{\left\vert\kern-0.25ex\left\vert\kern-0.25ex\left\vert #1
    \right\vert\kern-0.25ex\right\vert\kern-0.25ex\right\vert}}

\newcommand{\E}{\mathbb{E}}

\newcommand{\Th}{\boldsymbol{\theta}}
\newcommand{\Eta}{\boldsymbol{\eta}}

\newcommand{\SSigma}{\boldsymbol{\Sigma}}
\newcommand{\GGamma}{\boldsymbol{\Gamma}}
\newcommand{\LLambda}{\boldsymbol{\Lambda}}
\newcommand{\DDelta}{\boldsymbol{\Delta}}
\newcommand{\XXi}{\boldsymbol{\Xi}}

\DeclareMathOperator{\argmin}{argmin}

\DeclareMathOperator{\tr}{Tr}
\DeclareMathOperator{\cone}{cone}

\DeclareMathOperator{\Diag}{Diag}

\DeclareMathOperator{\conv}{conv}

\newcounter{exampleI}
{\theorembodyfont{\rmfamily} \theoremstyle{plain} }

\newcounter{exampleII}
{\theorembodyfont{\rmfamily} \theoremstyle{plain} }

\newcounter{exampleIII}
{\theorembodyfont{\rmfamily} \theoremstyle{plain} }

{\theorembodyfont{\rmfamily} }
{\theorembodyfont{\rmfamily} \newtheorem{defn}{Definition}}
\newtheorem{theo}{Theorem}

\newtheorem{lemm}{Lemma}
\newtheorem{corr}{Corollary}

\newcommand{\proof}{\noindent{\itshape Proof:}\hspace*{1em}}

\newcommand{\qed}{\nolinebreak[1]~~~\hspace*{\fill} \rule{5pt}{5pt}\vspace*{\parskip}\vspace*{1ex}}

\newcommand {\commentout}[1] {}

\def\ints{{{\rm Z} \kern -.35em {\rm Z} }}  
\def\smallints{{{\rm Z} \kern -.3em {\rm Z} }}  
\def\pints{{{\rm I} \kern -.15em {\rm N} }}      
\newcommand{\reals}{\mathbb R}

\def\cplx{{{\rm I} \kern -.45em {\rm C} }}       
\def\l2{\rm {\mathcal L}^{2}(\reals)}            

\newtheorem{nad}{Notation and Definitions}[section]

\newtheorem{theorem}{Theorem}[section]
\newtheorem{lemma}[theorem]{Lemma}

\newtheorem{proposition}{Proposition}[section]

\newcommand{\be}{\begin{eqnarray}}
\newcommand{\ee}{\end{eqnarray}}
\newcommand{\bea}{\begin{eqnarray}}
\newcommand{\eea}{\end{eqnarray}}
\newcommand{\beaa}{\begin{eqnarray*}}
\newcommand{\eeaa}{\end{eqnarray*}}
\newcommand{\bnad}{\begin{nad}}
\newcommand{\enad}{\end{nad}}

\newcommand{\diam}{{\rm diam\,}}

\title{Alternating Estimation for Structured High-Dimensional Multi-Response Models}
\date{\today}
\author{Sheng Chen \qquad \qquad Arindam Banerjee \vspace*{2mm}
\\
\{shengc,banerjee@cs.umn.edu\}\vspace*{2mm}\\
Department of Computer Science \& Engineering\\
University of Minnesota, Twin Cities}

\begin{document}

\maketitle

\begin{abstract}
  We consider learning high-dimensional multi-response linear models with structured parameters. By exploiting the noise correlations among responses, we propose an alternating estimation (AltEst) procedure to estimate the model parameters based on the generalized Dantzig selector. Under suitable sample size and resampling assumptions, we show that the error of the estimates generated by AltEst, with high probability, converges linearly to certain minimum achievable level, which can be tersely expressed by a few geometric measures, such as Gaussian width of sets related to the parameter structure. To the best of our knowledge, this is the first non-asymptotic statistical guarantee for such AltEst-type algorithm applied to estimation problem with general structures.  
\end{abstract}

\section{Introduction}
\label{sec:intro}
Multi-response (a.k.a. multivariate) linear models \cite{brfr97,izen08} have found numerous applications in real-world problems, e.g. expression quantitative trait loci (eQTL) 
mapping in computational biology \cite{kixi12}, land surface temperature prediction in climate informatics \cite{gdcs14}, neural semantic basis discovery in cognitive science \cite{lipz09}, etc. Unlike simple linear model where each response is a scalar, one obtains a \emph{response vector} at each observation in multi-response model, given as a (noisy) linear combinations of predictors, and the parameter (i.e., coefficient vector) to learn can be either response-specific (i.e., allowed to be different for every response), or shared by all responses. The multi-response model has been well studied under the context of the multi-task learning \cite{caru97}, where each response is coined as a \emph{task}. In recent years, the multi-task learning literature have largely focused on exploring the parameter structure across tasks via convex formulations \cite{evpo04,arep08,jsrr10}, which is not central to our current work. Another emphasis area in multi-response modeling is centered around the exploitation of the noise correlation among different responses \cite{rolz10,soki12,leli12}, instead of assuming that the noise is independent for each response. To be specific, we consider the following multi-response linear models with $m$ real-valued outputs,
\begin{gather}
\label{eq:mt_model}
\y_i = \bX_i \Th^* + \Eta_i, \quad \ \ \Eta_i \sim \cN(\mathbf{0}, \boldsymbol{\Sigma}_{*}) ~,
\end{gather}
where $\y_i \in \R^m$ is the response vector, $\bX_i \in \R^{m\times p}$ consists of $m$ $p$-dimensional feature vectors, and $\Eta_i \in \R^m$ is a noise vector sampled from a multivariate zero-mean Gaussian distribution with covariance $\boldsymbol{\Sigma}_{*}$. For simplicity, we assume $\Diag(\SSigma_*) = \bI_{m \times m}$ throughout the paper. The $m$ responses share the same underlying parameter $\Th^* \in \R^p$, which corresponds to the so-called \emph{pooled model} \cite{gree03}. In fact, this seemingly restrictive setting is general enough to encompass the model with response-specific parameters, which can be realized by block-diagonalizing rows of $\bX_i$ and stacking all coefficient vectors into a ``long'' vector.
Under the assumption of correlated noise, the true noise covariance structure $\boldsymbol{\Sigma}_*$ is usually unknown. Therefore it is typically required to estimate the parameter $\Th^*$ along with the covariance $\SSigma_*$.
In practice, we observe $n$ data points, denoted by $\cD = \{(\bX_i, \y_i)\}_{i=1}^n$, and the maximum likelihood estimator (MLE) is simply as follows,
\begin{align}
\label{eq:mle}
\left(\hat{\Th}_{\text{MLE}},  \hat{\SSigma}_{\text{MLE}}\right) 
= \underset{\substack{\Th \in \R^p, \ \SSigma \succeq 0}}{\argmin}  \ \frac{1}{2} \log \left| \SSigma \right|  + \frac{1}{2n} \sum_{i=1}^n \left\| \boldsymbol{\Sigma}^{-\frac{1}{2}} (\y_i - \bX_i \Th) \right\|_2^2
\end{align}
Although being convex w.r.t. either $\Th$ or $\SSigma$ when the other is fixed, the optimization problem associated with the MLE is jointly \emph{non-convex} for $\Th$ and $\SSigma$. A popular approach to dealing with such problem is \emph{alternating minimization} (AltMin), i.e., alternately solving for $\Th$ (and $\SSigma$) while keeping $\SSigma$ (and $\Th$) fixed. The AltMin algorithm for \eqref{eq:mle} iteratively performs two simple steps, solving least squares for $\Th$ and computing empirical noise covariance for $\SSigma$. Recent work \cite{jate15} has established the non-asymptotic error bound of this approach for \eqref{eq:mle} with a bit extension to sparse parameter setting using iterative hard thresholding method \cite{jatk14}. Previous works \cite{rolz10,leli12,rakd12} also considered the regularized MLE approaches for multi-response models with sparse parameters, which are solved by AltMin-type algorithms as well. Unfortunately, \emph{none} of those works provide \emph{finite-sample} statistical guarantees for their algorithms. AltMin technique has also been applied to many other problems, such as matrix completion \cite{jans13}, sparse coding \cite{aajn13}, and mixed linear regression \cite{yics14}, with provable performance guarantees. Despite the success of AltMin, most existing works are focused on recovering unstructured sparse or low-rank parameters, with little attention paid to general structures, e.g., overlapping sparsity \cite{jaov09}, hierarchical sparsity \cite{jmob11}, $k$-support sparsity \cite{arfs12}, etc.

In this paper, we focus on the multi-response linear model in high-dimensional setting, i.e., sample size $n$ is smaller than the problem dimension $p$, and the coefficient vector $\Th^*$ is assumed to possess a general low-complexity structure, which can be essentially captured by certain \emph{norm} $\|\cdot\|$ \cite{bjmo11}. Structured estimation using norm regularization/minimization has been extensively studied for simple linear models over the past decade, and recent advances manage to characterize the estimation error for convex approaches including Lasso-type (regularized) \cite{tibs96,nrwy12,bcfs14} and Dantzig-type (constrained) estimator \cite{birt09,chcb14}, via a few simple \emph{geometric measures}, e.g., Gaussian width \cite{gord85,crpw12} and restricted norm compatibility \cite{nrwy12,chcb14}.
Here we propose an \emph{alternating estimation} (AltEst) procedure for finding the true parameters, which essentially alternates between estimating $\Th$ through the generalized Dantzig selector (GDS) \cite{chcb14} using norm $\|\cdot\|$ and computing the approximated empirical noise covariance for $\SSigma$. Our analysis puts no restriction on what the norm can be, thus the AltEst framework is applicable to general structures. In contrast to AltMin, our AltEst procedure \emph{cannot} be casted as a minimization of some joint objective function for $\Th$ and $\SSigma$, thus is conceptually more general than AltMin. For the proposed  AltEst, we provide the statistical guarantees for the iterate $\hat{\Th}_t$ with the \emph{resampling} assumption (see Section \ref{sec:alt_est}), which may justify the applicability of AltEst technique to other problems without joint objectives for two parameters. Specifically, we show that with overwhelming probability, the estimation error $\|\hat{\Th}_t - \Th^*\|_2$ for generally structured $\Th^*$ converges \emph{linearly} to a \emph{minimum achievable error} given sub-Gaussian data under moderate sample size. With a straightforward intuition, this minimum achievable error can be tersely expressed by the aforementioned geometric measures which simply depend on the structure of $\Th^*$. Moreover, our analysis implies the error bound for single response high-dimensional models as a by-product \cite{chcb14}. Note that the analysis in \cite{jate15} focuses on the expected prediction error $\E [\SSigma_*^{-1 / 2} \bX (\hat{\Th}_t - \Th^*) ]$ for unstructured $\Th^*$, which is related but different from our $\|\hat{\Th}_t - \Th^*\|_2$ for generally structured $\Th^*$. Compared with the error bound derived for unstructured $\Th^*$ in \cite{jate15}, our result also yields better dependency on sample size by removing the $\log n$ factor, which seems unnatural to appear.

The rest of the paper is organized as follows. We elaborate our AltEst algorithm for high-dimensional multi-response linear models in Section \ref{sec:alt_est}, along with the resampling assumption. In Section \ref{sec:analysis}, we present the statistical guarantees for AltEst. We provide experimental results in Section \ref{sec:exp} to support our theoretical development. All proofs are deferred to the supplement.

\section{Alternating Estimation for High-Dimensional Multi-Response Models}
\label{sec:alt_est}
Given the high-dimensional setting for \eqref{eq:mt_model}, it is natural to consider the regularized MLE for \eqref{eq:mt_model} by adding the norm $\|\cdot\|$ to \eqref{eq:mle}, which captures the structural information of $\Th^*$ in \eqref{eq:mt_model},
\begin{align}
\label{eq:mle_re}
\begin{split}
\left(\hat{\Th}, \ \hat{\SSigma}\right) 
&= \underset{\Th \in \R^p, \ \SSigma \succeq 0}{\argmin} \  \frac{1}{2} \log \left| \SSigma \right|  + \frac{1}{2n} \sum_{i=1}^n \left\| \boldsymbol{\Sigma}^{-\frac{1}{2}} (\y_i - \bX_i \Th) \right\|_2^2 + \gamma_n \|\Th\| ~,
\end{split}
\end{align}
where $\gamma_n$ is a tuning parameter. Using AltMin the update of \eqref{eq:mle_re} can be given as
\begin{align}
\label{eq:mt_lasso}
& \hat{\Th}_{t} = \underset{\R^{p}}{\argmin} \ \frac{1}{2n} \sum_{i=1}^n \left\| \hat{\boldsymbol{\Sigma}}_{t-1}^{-\frac{1}{2}} (\y_i - \bX_i \Th) \right\|_2^2 + \gamma_n \|\Th\| \\
\label{eq:est_cov}
& \hat{\SSigma}_{t} = \frac{1}{n} \sum_{i=1}^n \left(\y_i - \bX_i \hat{\Th}_{t}\right) \left(\y_i - \bX_i \hat{\Th}_{t} \right)^T
\end{align}
The update of $\hat{\Th}_{t}$ is basically solving a regularized least squares problem, and the new $\hat{\SSigma}_{t}$ is obtained by computing the approximated empirical covariance of the residues evaluated at $\hat{\Th}_{t}$. In this work, we consider an alternative
to \eqref{eq:mt_lasso}, the generalized Dantzig selector (GDS) \cite{chcb14}, which is given by
\beq
\label{eq:mt_gds}
\hat{\Th}_{t} = \underset{\Th \in \R^p}{\argmin} \ \|\Th\| \quad \text{s.t.} \quad \left\| \frac{1}{n} \sum_{i=1}^n \bX^T_i \hat{\SSigma}_{t-1}^{-1} (\bX_i \Th - \y_i) \right\|_* \leq \gamma_n ~,
\eeq
where $\|\cdot\|_*$ is the \emph{dual norm} of $\|\cdot\|$. Compared with \eqref{eq:mt_lasso}, GDS has nicer geometrical properties, which is favored in the statistical analysis. More importantly, since iteratively solving \eqref{eq:mt_gds} followed by covariance estimation \eqref{eq:est_cov} no longer minimizes a specific objective function jointly, the updates go beyond the scope of AltMin, leading to our broader alternating estimation (AltEst) framework, i.e., alternately estimating one parameter by suitable approaches while keeping the other fixed. For the ease of exposition, we focus on the $m \leq n$ scenario, so that $\hat{\SSigma}_{t}$ can be easily computed in closed form as shown in \eqref{eq:est_cov}. When $m > n$ and $\SSigma_*^{-1}$ is sparse, it is beneficial to directly estimate $\SSigma_*^{-1}$ using more advanced estimators \cite{frht08,call11}. Especially the CLIME estimator \cite{call11} enjoys certain desirable properties, which fits into our AltEst framework but not AltMin, and
our AltEst analysis \emph{does not} rely on the particular estimator we use to estimate noise covariance or its inverse. The algorithmic details are given in Algorithm \ref{alg:alt_est}, for which it is worth noting that every iteration $t$ uses independent new samples, $\cD_{2t-1}$ and $\cD_{2t}$ in Step \ref{step:gds} and \ref{step:cov}, respectively. This assumption is known as \emph{resampling}, which facilitates the theoretical analysis by removing the statistical dependency between iterates. Several existing works benefit from such assumption when analyzing their AltMin-type algorithms \cite{jans13,nejs13,yics14}. Conceptually resampling can be implemented by partitioning the whole dataset into $T$ subsets, though it is unusual to do so in practice. 
Loosely speaking, AltEst (AltMin) with resampling is an approximation of the practical AltEst (AltMin) with a single dataset $\cD$ used by all iterations. For AltMin, attempts have been made to directly analyze its practical version without resampling, by studying the properties of the joint objective \cite{sulu15}, which come at the price of invoking highly sophisticated mathematical tools. This technique, however, might fail to work for AltEst since the procedure is not even associated with a joint objective. Therefore in the next section, we will leverage the resampling assumption to show that the estimation error of $\hat{\Th}_t$ generated by Algorithm \ref{alg:alt_est} will converge to a small value with high probability.
\begin{algorithm}[hbt!]
\renewcommand{\algorithmicrequire}{\textbf{Input:}}
\renewcommand{\algorithmicensure} {\textbf{Output:} }
\caption{Alternating Estimation with Resampling}
\label{alg:alt_est}
\begin{algorithmic}[1]
\REQUIRE Number of iterations $T$, Datasets $\cD_1 = \{(\bX_i, \y_i)\}_{i=1}^{n}$, 
$\ldots$ , $\cD_{2T} = \{(\bX_i, \y_i)\}_{i=(2T-1)n+1}^{2Tn}$ \\
\STATE Initialize $\hat{\SSigma}_0 = \bI_{m \times m}$
\FOR {$t$:= $1$ to $T$}
\STATE Solve the GDS \eqref{eq:mt_gds} for $\hat{\Th}_t$ using dataset $\cD_{2t-1}$
\label{step:gds}
\STATE Compute $\hat{\SSigma}_t$ according to \eqref{eq:est_cov} using dataset $\cD_{2t}$
\label{step:cov}
\ENDFOR
\RETURN $\hat{\Th}_T$
\end{algorithmic}
\end{algorithm}

\section{Statistical Guarantees for Alternating Estimation}
\label{sec:analysis}
In this section, we establish the statistical guarantees for our AltEst algorithm. The road map for the analysis is to first derive the error bounds separately for both coefficient vector and noise covariance estimation, and then combine them through AltEst procedure to show the error bound of $\hat{\Th}_t$. Throughout the analysis, the design $\bX$ is assumed to centered, i.e., $\E[\bX] = \mathbf{0}_{m \times p}$. $\lambda_{\max}(\cdot)$ and $\lambda_{\min}(\cdot)$ are used to denote the largest and smallest eigenvalue of a real symmetric matrix. Before presenting the main results, we provide some basic concepts and knowledge, which will be used in our analysis. First of all, we introduce the definition of sub-Gaussian matrix $\bX$.
\begin{defn}[Sub-Gaussian Matrix]
$\bX \in \R^{m \times p}$ is sub-Gaussian if the $\psi_2$-norm below is finite,
\beq
\label{eq:subgauss_mat}
\vertiii{\bX}_{\psi_2} = \sup_{\substack{\v \in \s^{p-1}, \ \u \in \s^{m-1}}} \ \vertiii{\v^T \GGamma_{\u}^{-\frac{1}{2}} \bX^T \u}_{\psi_2} \leq \kappa < + \infty ~,
\eeq
where $\GGamma_{\u} = \E[\bX^T \u \u^T \bX]$. Further we assume there exist constants $\mu_{\min}$ and $\mu_{\max}$ such that
\beq
\label{eq:mu_max_min}
0 < \mu_{\min} \leq \lambda_{\min}(\GGamma_{\u}) \leq \lambda_{\max}(\GGamma_{\u}) \leq  \mu_{\max} < + \infty ~, \quad  \forall \ \u \in \s^{m-1}
\eeq
\end{defn}
The definition \eqref{eq:subgauss_mat} is also used in earlier work \cite{jate15}, which assumes the left end of \eqref{eq:mu_max_min} implicitly. Lemma \ref{lem:subgauss_exp} gives an example of sub-Gaussian $\bX$, showing that condition \eqref{eq:subgauss_mat} and \eqref{eq:mu_max_min} are reasonable.
\begin{lemm}
\label{lem:subgauss_exp}
Assume that $\bX \in \R^{m \times p}$ has dependent anisotropic rows such that $\bX = \XXi^{\frac{1}{2}} \tilde{\bX} \LLambda^{\frac{1}{2}}$, where $\XXi \in \R^{m \times m}$ encodes the dependency between rows, $\tilde{\bX} \in \R^{m \times p}$ has independent isotropic rows, and $\LLambda \in \R^{p \times p}$ introduces the anisotropy. In this setting, if each row of $\tilde{\bX}$ satisfies $\vertiii{\tilde{\x}_i}_{\psi_2} \leq \tilde{\kappa}$, then condition \eqref{eq:subgauss_mat} and \eqref{eq:mu_max_min} hold with $\kappa = C \tilde{\kappa}$, $\mu_{\min} = \lambda_{\min}(\XXi) \lambda_{\min}(\LLambda)$, and $\mu_{\max} = \lambda_{\max}(\XXi) \lambda_{\max}(\LLambda)$.
\end{lemm}

The recovery guarantee of GDS relies on an important notion called \emph{restricted eigenvalue} (RE). In multi-response setting, it is defined jointly for designs $\bX_i$ and a noise covariance $\SSigma$ as follows.
\begin{defn}[Restricted Eigenvalue Condition]
The designs $\bX_1, \bX_2, \ldots, \bX_n$ and the covariance $\SSigma$ together satisfy the restricted eigenvalue condition for set $\cA \in \s^{p-1}$ with parameter $\alpha > 0$, if
\beq
\label{eq:re_def}
\inf_{\v \in \cA} \ \v^T \left( \frac{1}{n} \sum_{i=1}^n \bX_i^T \SSigma^{-1} \bX_i \right) \v \ \geq \ \alpha ~.
\eeq
\end{defn}
Apart from RE condition, the analysis of GDS is carried out on the premise that tuning parameter $\gamma_n$ is suitably selected, which we define as ``admissible''.
\begin{defn}[Admissible Tuning Parameter]
The $\gamma_n$ for GDS \eqref{eq:mt_gds} is said to be \emph{admissible} if $\gamma_n$ is chosen such that $\Th^*$ belongs to the constraint set, i.e.,
\beq
\label{eq:ad_gamma}
\left\| \frac{1}{n} \sum_{i=1}^n \bX^T_i \SSigma^{-1} (\bX_i \Th^* - \y_i) \right\|_* = \left\| \frac{1}{n} \sum_{i=1}^n \bX^T_i \SSigma^{-1} \Eta_i \right\|_* \leq \gamma_n
\eeq
\end{defn}
For structured estimation, one also needs to characterize the structural complexity of $\Th^*$, and an appropriate choice is the \emph{Gaussian width} \cite{gord85} defined as follows.
\begin{defn}[Gaussian Width]
For any set $\cA \subseteq \R^p$, its Gaussian width is given by $w(\cA) = \E \left[ \sup_{\u \in \cA} \ \langle \u, \g \rangle \right]$,
where $\g \sim \cN(\mathbf{0}, \bI_{p \times p})$ is a standard Gaussian random vector.
\end{defn}
In the analysis, the set $\cA$ of our interests typically relies on the structure of $\Th^*$. Previously Gaussian width has been applied to statistical analyses for various problems \cite{crpw12,bcfs14,trop15}, and recent works \cite{rarn12,chba15} show that Gaussian width is computable for many structures. For the rest of the paper, we use $C, C_0, C_1$ and so on to denote universal constants, which are different from context to context.

\subsection{Estimation of Coefficient Vector}
\label{sec:gds_analysis}
In this subsection, we focus on estimating $\Th^*$, i.e., Step \ref{step:gds} of Algorithm \ref{alg:alt_est}, using GDS of the form,
\beq
\label{eq:gds}
\hat{\Th} = \underset{\Th \in \R^p}{\argmin} \ \|\Th\| \quad \text{s.t.} \quad \left\| \frac{1}{n} \sum_{i=1}^n \bX^T_i \SSigma^{-1} (\bX_i \Th - \y_i) \right\|_* \leq \gamma_n ~,
\eeq
where $\SSigma$ is an arbitrary but fixed input noise covariance matrix. The following lemma shows a \emph{deterministic} error bound for $\hat{\Th}$ under the RE condition and admissible $\gamma_n$ defined in \eqref{eq:re_def} and \eqref{eq:ad_gamma}.
\begin{lemm}
\label{lem:gds}
Suppose the RE condition \eqref{eq:re_def} is satisfied by $\bX_1, \ldots, \bX_n$ and $\SSigma$ with $\alpha > 0$ for the set $\cA\left(\Th^*\right) = \cone \left\{ \v \ | \ \left\|\Th^* + \v \right\| \leq \left\|\Th^*\right\| \ \right\} \cap \s^{p-1}$. If $\gamma_n$ is admissible,  $\hat{\Th}$ in \eqref{eq:gds} satisfies
\beq
\left\|\hat{\Th} - \Th^*\right\|_2 \leq 2\Psi(\Th^*) \cdot \frac{\gamma_n}{\alpha} ~,
\eeq
in which $\Psi(\Th^*)$ is the restricted norm compatibility defined as $\Psi(\Th^*) = \sup_{\v \in \cA\left(\Th^*\right)} \frac{\|\v\|}{\|\v\|_2}$.
\end{lemm}

From Lemma \ref{lem:gds}, we can find that the $L_2$-norm error is mainly determined by three quantities--$\Psi(\Th^*)$, $\gamma_n$ and $\alpha$. The restricted norm compatibility $\Psi(\Th^*)$ is purely hinged on the geometrical structure of $\Th^*$ and $\|\cdot\|$, thus involving no randomness. On the contrary, $\gamma_n$ and $\alpha$ need to satisfy their own conditions, which are bound to deal with random $\bX_i$ and $\Eta_i$. The set $\cA(\Th^*)$ involved in RE condition and restricted norm compatibility has relatively simple structure, which will favor the derivation of error bound for varieties of norms \cite{chba15}. If RE condition fails to hold, i.e. $\alpha = 0$, the error bound is meaningless. Though the error is proportional to the user-specified $\gamma_n$, assigning arbitrarily small value to $\gamma_n$ may not be admissible.  Hence, in order to further derive the recovery guarantees for GDS, we need to verify RE condition and find the smallest admissible value of $\gamma_n$.

\textbf{Restricted Eigenvalue Condition:} Firstly the following lemma characterizes the relation between the expectation and empirical mean of $\bX^T \SSigma^{-1} \bX$.
\begin{lemm}
\label{lem:re}
Given sub-Gaussian $\bX \in \R^{m \times p}$ with its i.i.d. copies $\bX_1, \ldots, \bX_n$, and covariance $\SSigma \in \R^{m \times m}$ with eigenvectors $\u_1, \ldots, \u_m$, let $\GGamma = \E [\bX^T \SSigma^{-1} \bX]$ and $\hat{\GGamma} = \frac{1}{n} \sum_{i=1}^n \bX_i^T \SSigma^{-1} \bX_i$. Define the following set for $\cA \subseteq \s^{p-1}$ and each $\GGamma_j = \E[ \bX^T \u_j \u_j^T \bX]$,
\begin{align}
\label{eq:re_set}
\cA_{\GGamma_j} = \left\{ \v \in \s^{p-1} \ | \ \GGamma^{-\frac{1}{2}}_j \v \in \cone(\cA) \right\} ~.
\end{align}
If $n \geq C_1 \kappa^4 \cdot \max_j \left\{w^2(\cA_{\GGamma_j})\right\}$, with probability at least $1 - m \exp(-C_2 n / \kappa^4)$, we have
\beq
\v^T \hat{\GGamma} \v \geq \frac{1}{2} \v^T \GGamma \v,  \quad \forall \ \v \in \cA ~.
\eeq
\end{lemm}

Instead of $w(\cA_{\GGamma_j})$, ideally we want the condition above on $n$ to be characterized by  $w(\cA)$, which can be easier to compute in general. The next lemma accomplishes this goal.
\begin{lemm}
\label{lem:width_rel}
Let $\kappa_0$ be the $\psi_2$-norm of standard Gaussian random vector and $\GGamma_{\u} = \E [\bX^T \u \u^T \bX]$, where $\u \in \s^{m-1}$ is fixed. Given $\cA \subseteq \s^{p-1}$, define the set $\cA_{\GGamma_{\u}}$ according to \eqref{eq:re_set}. Then we have
\beq
w(\cA_{\GGamma_{\u}}) \leq C \kappa_0 \sqrt{\mu_{\max} / \mu_{\min}} \cdot \left( w(\cA) + 3 \right) ~,
\eeq
\end{lemm}

Lemma \ref{lem:width_rel} implies that the Gaussian width $w(\cA_{\GGamma_{j}})$ appearing in Lemma \ref{lem:re} is at the same order of $w(\cA)$. Putting Lemma \ref{lem:re} and \ref{lem:width_rel} together, we can obtain the RE condition for the analysis of GDS.
\begin{corr}
\label{cor:re}
Under the notations of Lemma \ref{lem:re} and \ref{lem:width_rel}, if $n \geq C_1 \kappa_0^2 \kappa^4 \cdot  \frac{\mu_{\max}}{\mu_{\min}} \cdot (w(\cA) + 3)^2$, then the following inequality holds for all $\v \in \cA \subseteq \s^{p-1}$ with probability at least $1 - m \exp(-C_2 n / \kappa^4)$,
\beq
\v^T \hat{\GGamma} \v \geq \frac{\mu_{\min}}{2} \cdot \tr(\SSigma^{-1})
\eeq
\end{corr}

\textbf{Admissible Tuning Parameter:} Finding the admissible $\gamma_n$ amounts to estimating the value of $\|\frac{1}{n} \sum_{i=1}^n \bX_i^T \SSigma^{-1} \Eta_i \|_*$ in \eqref{eq:ad_gamma}, which involves random $\bX_i$ and $\Eta_i$. The next lemma establishes a high-probability bound for this quantity, which can be viewed as the smallest ``safe'' choice of $\gamma_n$.
\begin{lemm}
\label{lem:gamma_n}
Assume that $\bX_i$ is sub-Gaussian and $\Eta_i \sim \cN(\mathbf{0}, \SSigma_*)$. The following inequality holds with probability at least $1 - \exp\left(-\frac{n \tau^2}{2}\right) - C_2\exp(-\frac{C_1^2 w^2(\cB)}{4 \rho^2})$
\beq
\left\|\frac{1}{n} \sum_{i=1}^n \bX_i^T \SSigma^{-1} \Eta_i  \right\|_* \leq \frac{C \kappa \sqrt{\mu_{\max}} }{\sqrt{n}}  \cdot \sqrt{ \tr\left(\SSigma^{-1}  \SSigma_* \SSigma^{-1}\right)}  \cdot w(\cB) ~,
\eeq
where $\cB$ denotes the unit ball of norm $\|\cdot\|$, $\rho = \sup_{\v \in \cB} \|\v\|_2$, and $\tau = \|\SSigma^{-1} \SSigma_*^{\frac{1}{2}}\|_{F} / \|\SSigma^{-1} \SSigma_*^{\frac{1}{2}}\|_{2}$.
\end{lemm}

\textbf{Estimation Error of GDS:} Building on Corollary \ref{cor:re}, Lemma \ref{lem:gds} and \ref{lem:gamma_n}, the theorem below characterizes the estimation of GDS for the multi-response linear model.
\begin{theo}
\label{the:gds_bound}
Under the setting of Lemma \ref{lem:gamma_n}, if $n \geq C_1 \kappa_0^2 \kappa^4 \cdot  \frac{\mu_{\max}}{\mu_{\min}} \cdot (w(\cA\left(\Th^*\right)) + 3)^2$, and $\gamma_n$ is set to $C_2 \kappa \sqrt{\frac{\mu_{\max} \tr\left(\SSigma^{-1}  \SSigma_* \SSigma^{-1}\right)}{n}}  \cdot w(\cB)$, the estimation error of $\hat{\Th}$ given by \eqref{eq:gds} satisfies
\beq
\|\hat{\Th} - \Th^*\|_2 \leq  C\kappa \sqrt{\frac{\mu_{\max}}{\mu_{\min}^2}}  \cdot \frac{\sqrt{ \tr\left(\SSigma^{-1}  \SSigma_* \SSigma^{-1}\right)}}{\tr\left(\SSigma^{-1}\right)} \cdot \frac{\Psi(\Th^*) \cdot w(\cB)}{\sqrt{n}}  ~,
\eeq
with probability at least $1 - m \exp\left(-\frac{C_3 n}{\kappa^4}\right) - \exp\left(-\frac{n \tau^2}{2}\right) - C_4\exp(-\frac{C_5^2 w^2(\cB)}{4 \rho^2})$.
\end{theo}
\textbf{Remark:} We can see from the theorem above that the noise covariance $\SSigma$ input to GDS plays a role in the error bound through the multiplicative factor $\xi(\SSigma) = \sqrt{ \tr\left(\SSigma^{-1}  \SSigma_* \SSigma^{-1}\right)} / \tr\left(\SSigma^{-1}\right)$. By taking the derivative of $\xi^2(\SSigma)$ w.r.t. $\SSigma^{-1}$ and setting it to $\mathbf{0}$, we have
\begin{align*}
\frac{\partial \xi^2(\SSigma)}{\partial \SSigma^{-1}} = \frac{2\tr^2\left(\SSigma^{-1}\right) \SSigma_* \SSigma^{-1} - 2\tr\left(\SSigma^{-1}\right) \tr\left(\SSigma^{-1}  \SSigma_* \SSigma^{-1}\right) \cdot \bI_{m \times m}}{\tr^4\left(\SSigma^{-1}\right) } = \mathbf{0}
\end{align*}
Then we can verify that $\SSigma = \SSigma_*$ is the solution to the equation above, and thus is the minimizer of $\xi(\SSigma)$ with $\xi(\SSigma_*) = 1 / \sqrt{\tr(\SSigma^{-1}_* )}$. This calculation confirms that multi-response regression could benefit from taking into account the noise covariance, and the best performance is achieved when $\SSigma_*$ is known. If we perform ordinary GDS by setting $\SSigma = \bI_{m \times m}$, then $\xi(\SSigma) = 1 /\sqrt{m}$. Therefore using $\SSigma_*$ will reduce the error by a factor of $\sqrt{m / \tr(\SSigma_*^{-1})}$, compared with ordinary GDS.

One simple structure of $\Th^*$ to consider for Theorem \ref{the:gds_bound} is the sparsity encoded by $L_1$ norm. Given $s$-sparse $\Th^*$, it follows from previous results \cite{nrwy12,crpw12} that $\Psi(\Th^*) = O(\sqrt{s})$, $w(\cA(\Th^*)) = O(\sqrt{s \log p})$ and $w(\cB) = O(\sqrt{\log p})$. Therefore if $n \geq O(s \log p)$, then with high probability we have
\begin{align}
\|\hat{\Th} - \Th^*\|_2 \leq O\left(\xi(\SSigma)\cdot \sqrt{\frac{s \log p}{n}}\right)
\end{align}
\textbf{Implications for Simple Linear Models:}
Our general result in multi-response scenario implies some existing results for simple linear models. If we set $n = 1$ and $\SSigma = \SSigma_* = \bI_{m \times m}$, i.e., only one data point $(\bX, \y)$ is observed and the noise is independent for each response, the GDS is reduced to
\beq
\label{eq:gds_single}
\hat{\Th}_{\text{sg}} =  \underset{\Th \in \R^p}{\argmin} \ \|\Th\| \quad \text{s.t.} \quad \left\| \bX^T (\bX \Th - \y) \right\|_* \leq \gamma ~,
\eeq
which exactly matches that in \cite{chcb14}. To bound its estimation error, we need $\bX$ to be more structured beyond the sub-Gaussianity. Essentially we consider the model of $\bX$ in Lemma \ref{lem:subgauss_exp}, where rows of $\tilde{\bX}$ are additionally assumed to be identical. For such $\bX$, a specialized RE condition is as follows.
\begin{lemm}
\label{lem:re_special}
Assume $\bX$ is defined as in Lemma \ref{lem:subgauss_exp} such that $\bX = \XXi^{\frac{1}{2}} \tilde{\bX} \LLambda^{\frac{1}{2}}$, and rows of $\tilde{\bX}$ are i.i.d. with $\vertiii{\tilde{\x}_j} \leq \tilde{\kappa}$. If $m n \geq C_1 \kappa_0^2 \tilde{\kappa}^4 \cdot \frac{\lambda_{\max}(\XXi)\lambda_{\max}(\LLambda)}{\lambda_{\min}(\XXi)\lambda_{\min}(\LLambda)} \cdot \left( w(\cA) + 3 \right)^2$, with probability at least $1 - \exp(-C_2 mn / \tilde{\kappa}^4)$, the following inequality is satisfied by all $\v \in \cA \subseteq \s^{p-1}$,
\beq
\label{eq:re_special}
\v^T \hat{\GGamma} \v \ \geq \ \frac{m}{2} \cdot \lambda_{\min}\left( \XXi^{\frac{1}{2}} \SSigma^{-1} \XXi^{\frac{1}{2}} \right) \cdot \lambda_{\min}\left(\LLambda\right) ~.
\eeq
\end{lemm}
\textbf{Remark:} Lemma \ref{lem:re_special} characterizes the RE condition for a class of specifically structured design $\bX$. If we specialize the general RE condition in Corollary \ref{cor:re} for this setting, $\bX = \XXi^{\frac{1}{2}} \tilde{\bX} \LLambda^{\frac{1}{2}}$, it becomes
\begin{gather*}
n \geq C_1 \kappa_0^2 \tilde{\kappa}^4  \frac{\lambda_{\max}(\XXi)\lambda_{\max}(\LLambda)}{\lambda_{\min}(\XXi)\lambda_{\min}(\LLambda)} (w(\cA) + 3)^2  \xRightarrow{\substack{\text{with probability} \  1 - \\ m\exp(-C_2 n / \tilde{\kappa}^4) \\ \ }}
\v^T \hat{\GGamma} \v \geq \frac{\lambda_{\min}(\XXi)\lambda_{\min}(\LLambda)}{2} \tr(\SSigma^{-1}) 
\end{gather*}
Comparing the general result above with Lemma \ref{lem:re_special}, there are two striking differences. Firstly, Lemma \ref{lem:re_special}  requires the same sample size of $mn$ rather than $n$, which improves the general one. Secondly, 
\eqref{eq:re_special} holds with much higher probability $1- \exp(-C_2 mn / \tilde{\kappa}^4)$ instead of $1 - m \exp(-C_2 n / \tilde{\kappa}^4)$.

Given this specialized RE condition, we have the recovery guarantees of GDS for simple linear models, which encompass the settings discussed in \cite{bcfs14,chcb14} as special cases.
\begin{corr}
\label{cor:single}
Suppose $\y = \bX \Th^* + \Eta \in \R^m$, where $\bX$ is described as in Lemma \ref{lem:re_special}, and $\Eta \sim \cN(\mathbf{0}, \bI)$. With probability at least $1 - \exp\left(-\frac{m}{2}\right) - C_2\exp(-\frac{C_1^2 w^2(\cB)}{4 \rho^2}) - \exp(-C_3 m / \tilde{\kappa}^4)$, $\hat{\Th}_{\text{sg}}$ satisfies
\beq
\left\| \hat{\Th}_{\text{sg}} - \Th^* \right\|_2 \leq C \tilde{\kappa} \cdot \sqrt{ \frac{\lambda_{\max}(\XXi) \lambda_{\max}(\LLambda)}{\lambda^2_{\min}(\XXi) \lambda^2_{\min}(\LLambda)}} \cdot \frac{\Psi(\Th^*) \cdot w(\cB)}{\sqrt{m}}  ~,
\eeq
\end{corr}

\subsection{Estimation of Noise Covariance}
\label{sec:cov_analysis}
In this subsection, we consider the estimation of noise covariance $\SSigma_*$ given an arbitrary parameter vector $\Th$. When $m$ is small, we estimate $\SSigma_*$ by simply using the sample covariance
\vspace{-2mm}
\beq
\label{eq:cov}
\hat{\SSigma} = \frac{1}{n} \sum_{i=1}^n \left(\y_i - \bX_i \Th\right) \left(\y_i - \bX_i \Th \right)^T ~.
\eeq
Theorem \ref{the:est_cov} reveals the relation between $\hat{\SSigma}$ and $\SSigma_*$, which is sufficient for our AltEst analysis.
\vspace{-3mm}
\begin{theo}
\label{the:est_cov}
If $n \geq C^4 m \cdot \max \left\{ 4\left(\kappa_0 + \kappa \sqrt{\frac{\mu_{\max}}{{\lambda_{\min}\left( \SSigma_* \right)}}} \left\|\Th^* - \Th\right\|_2 \right)^4, \kappa^4 \left(\frac{\lambda_{\max}\left( \SSigma_* \right)\mu_{\max}}{\lambda_{\min}\left( \SSigma_* \right)\mu_{\min}} \right)^2 \right\}$ and $\bX_i$ is sub-Gaussian, with probability at least $1 - 2\exp(- C_1 m)$, $\hat{\SSigma}$ given by \eqref{eq:cov} satisfies
\begin{gather}
\label{eq:cov_max}
\lambda_{\max}\left(  \SSigma_*^{-\frac{1}{2}} \hat{\SSigma} \SSigma_*^{-\frac{1}{2}} \right) \ \leq \ 1 + C^2 \kappa_0^2 \sqrt{m / n} + \frac{2\mu_{\max} }{{\lambda_{\min}\left( \SSigma_* \right)}} \left\|\Th^* - \Th\right\|_2^2 \\
\label{eq:cov_min}
\lambda_{\min}\left(  \SSigma_*^{-\frac{1}{2}} \hat{\SSigma} \SSigma_*^{-\frac{1}{2}} \right) \ \geq \ 1 - C^2 \kappa_0^2 \sqrt{m / n}
\end{gather}
\end{theo}
\vspace{-3mm}
\textbf{Remark:} If $\hat{\SSigma} = \SSigma_*$, then $\lambda_{\max}(  \SSigma_*^{-\frac{1}{2}} \hat{\SSigma} \SSigma_*^{-\frac{1}{2}}) = \lambda_{\min}(  \SSigma_*^{-\frac{1}{2}} \hat{\SSigma} \SSigma_*^{-\frac{1}{2}}) = 1$. Hence $\hat{\SSigma}$ is nearly equal to $\SSigma_*$ when the upper and lower bounds \eqref{eq:cov_max} \eqref{eq:cov_min} are close to 1.  We would like to point out that there is nothing specific to the particular form of estimator \eqref{eq:cov}, which makes AltEst work. Similar results can be obtained for other methods that estimate the inverse covariance matrix $\SSigma_*^{-1}$ instead of $\SSigma_*$.  For instance, when $m < n$ and $\SSigma_*^{-1}$ is sparse, we can replace \eqref{eq:cov} with GLasso \cite{frht08} or CLIME \cite{call11}, and AltEst only requires the counterparts of \eqref{eq:cov_max} and \eqref{eq:cov_min} in order  to work.


\subsection{Error Bound for Alternating Estimation}
\label{sec:ae_analysis}
Section \ref{sec:gds_analysis} shows that the noise covariance in GDS affects the error bound by the factor $\xi(\SSigma)$. In order to bound the error of $\hat{\Th}_T$ given by AltEst, we need to further quantify how $\Th$ affects $\xi(\hat{\SSigma})$.
\begin{lemm}
\label{lem:factor_rel}
If $\hat{\SSigma}$ is given as \eqref{eq:cov} and the condition in Theorem \ref{the:est_cov} holds, then the inequality below holds with probability at least $1 - 2\exp(- C_1 m)$,
\beq
\xi\left(\hat{\SSigma}\right) \leq \xi\left(\SSigma_*\right) \cdot \left( 1 + 2C \kappa_0\left(\frac{m}{n} \right)^{\frac{1}{4}} + 2\sqrt{\frac{\mu_{\max}}{{\lambda_{\min}\left( \SSigma_* \right)}}} \left\|\Th^* - \Th\right\|_2 \right)
\eeq
\end{lemm}
\vspace{-3mm}
Based on Lemma \ref{lem:factor_rel}, the following theorem provides the error bound for $\hat{\Th}_T$ given by Algorithm \ref{alg:alt_est}.
\begin{theo}
\label{the:alt_est}
Let $e_{\text{orc}} = C_1\kappa \sqrt{\frac{\mu_{\max}}{\mu_{\min}^2}} \frac{\xi\left( \SSigma_*\right) \cdot \Psi(\Th^*) w(\cB)}{\sqrt{n}}$ and $e_{\text{min}} = e_{\text{orc}} \cdot \frac{ 1 + 2C \kappa_0\left(\frac{m}{n} \right)^{\frac{1}{4}}}{1 - 2 e_{\text{orc}} \sqrt{\frac{\mu_{\max}}{\lambda_{\min} ( \SSigma_* )}}}$. If $n \geq C^4 m \cdot \max \left\{ 4\left(\kappa_0 + \frac{C_1}{C^2} \sqrt{\frac{\lambda_{\min}\left( \SSigma_* \right)}{\lambda^2_{\max}\left( \SSigma_* \right)}} \frac{\Psi(\Th^*) w(\cB)}{m} \right)^4,  \kappa^4 \left(\frac{\lambda_{\max}\left( \SSigma_* \right)\mu_{\max}}{\lambda_{\min}\left( \SSigma_* \right)\mu_{\min}} \right)^2, \left( \frac{2 C_1  \kappa \mu_{\max}}{C^2 \mu_{\min}} \cdot \frac{\xi(\SSigma_*) \Psi(\Th^*) w(\cB)}{\sqrt{m\cdot \lambda_{\min}(\SSigma_*)}} \right)^2 \right\}$ and also satisfies the condition in Theorem \ref{the:gds_bound}, with high probability, the iterate $\hat{\Th}_T$ returned by Algorithm \ref{alg:alt_est} satisfies
\beq
\left\|\hat{\Th}_{T} - \Th^*\right\|_2 \ \leq \ e_{\text{min}} + \left(2 e_{\text{orc}} \sqrt{\frac{\mu_{\max}}{\lambda_{\min}\left( \SSigma_* \right)}} \right)^{T-1} \cdot \left( \left\|\hat{\Th}_{1} - \Th^*\right\|_2 - e_{\text{min}} \right)
\eeq
\end{theo}
\textbf{Remark:} The three lower bounds for $n$ inside curly braces correspond to three intuitive requirements. The first one guarantees that the covariance estimation is accurate enough, and the other two respectively ensure that the initial error of $\hat{\Th}_1$ and $e_{\text{orc}}$ are reasonably small , such that the subsequent errors can contract linearly. $e_{\text{orc}}$ is the estimation error incurred by the following oracle estimator,
\vspace{-1.3mm}
\beq
\label{eq:orc_gds}
\hat{\Th}_{\text{orc}} = \underset{\Th \in \R^p}{\argmin} \ \|\Th\| \quad \text{s.t.} \quad  \| \frac{1}{n} \sum_{i=1}^n \bX^T_i \SSigma_*^{-1} (\bX_i \Th - \y_i) \|_* \leq \gamma_n ~,
\eeq
which is impossible to implement in practice. On the other hand, $e_{\text{min}}$ is the minimum achievable error, which has an extra multiplicative factor compared with $e_{\text{orc}}$. The numerator of the factor compensates for the error of estimated noise covariance provided that $\Th = \Th^*$ is plugged in \eqref{eq:cov}, which merely depends on sample size. Since having $\Th = \Th^*$ is also unrealistic for \eqref{eq:cov}, the denominator further accounts for the ballpark difference between $\Th$ and $\Th^*$.

\section{Experiments}
\label{sec:exp}
\begin{figure*}[!hbt]
\centering
\vspace{-2.8mm}
\subfigure[Error for AltEst] {
    \label{fig:alt_est}
    \includegraphics[width=0.31\linewidth]{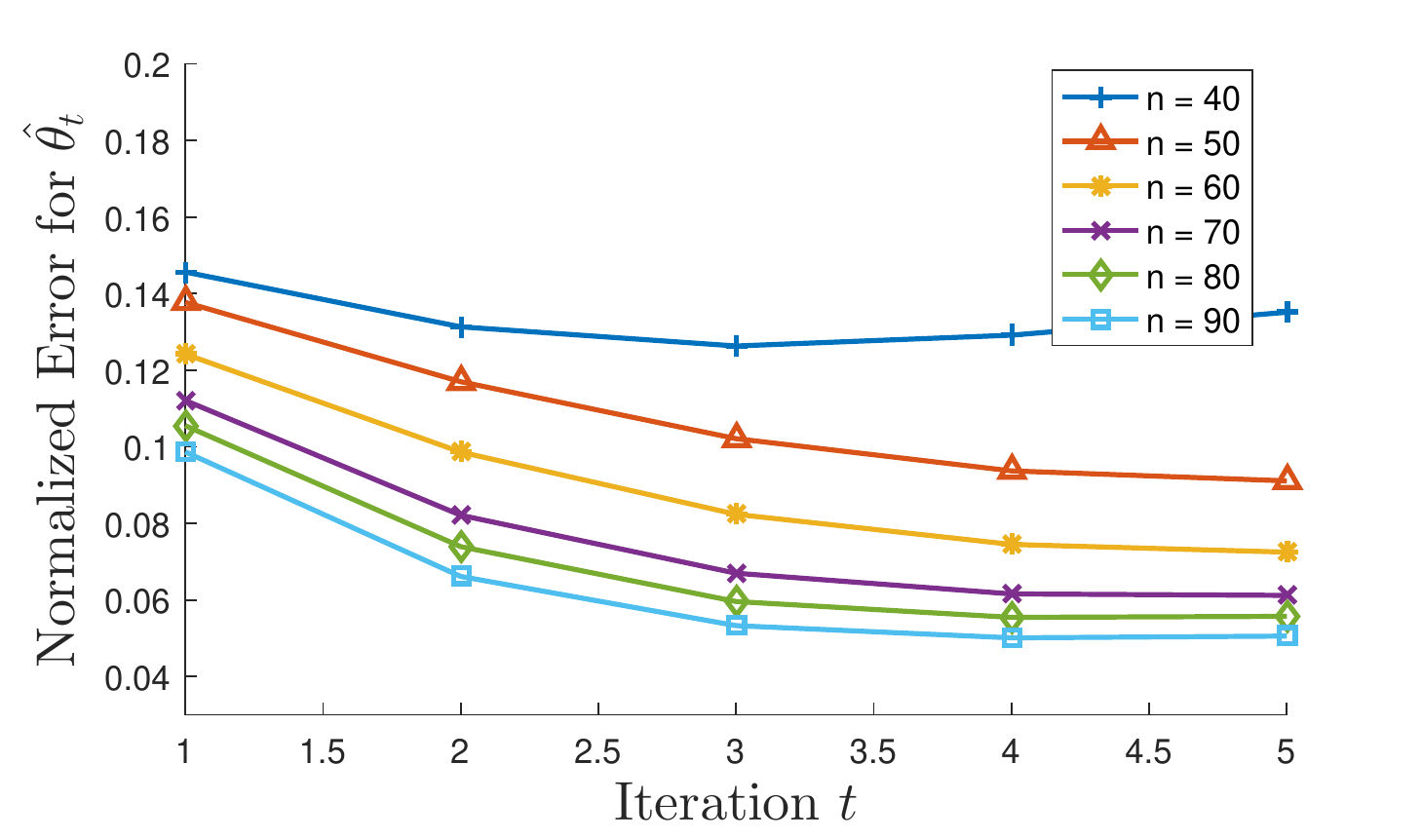}}
\begin{minipage}[b]{0.35in}

\end{minipage}
\subfigure[Error for Resampled AltEst] {
    \label{fig1:alt_est_rsp}
    \includegraphics[width=0.31\linewidth]{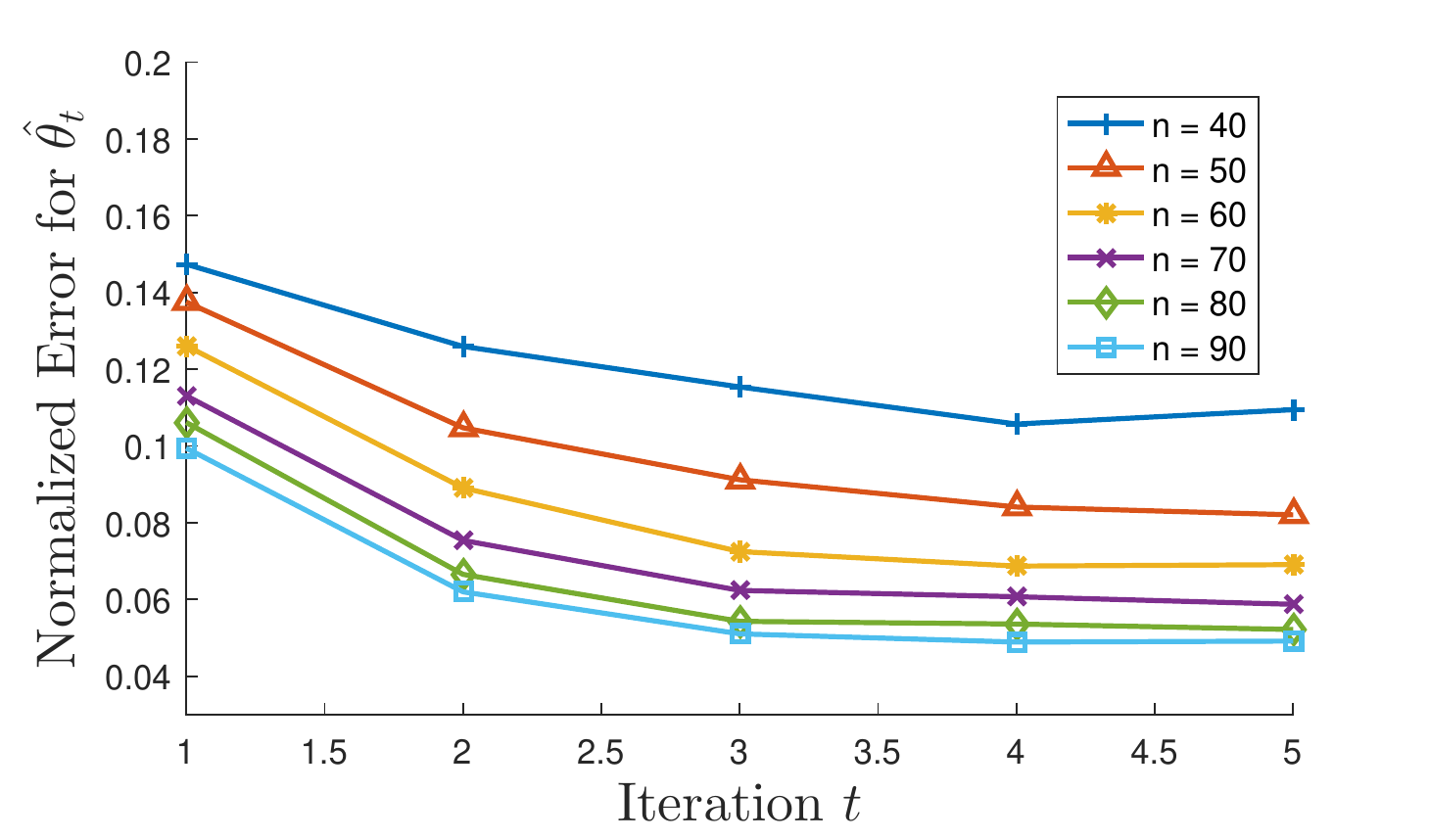}}
\begin{minipage}[b]{0.35in}

\end{minipage}
\subfigure[Comparison of Estimators] {
    \label{fig1:err_cmpr}
    \includegraphics[width=0.31\linewidth]{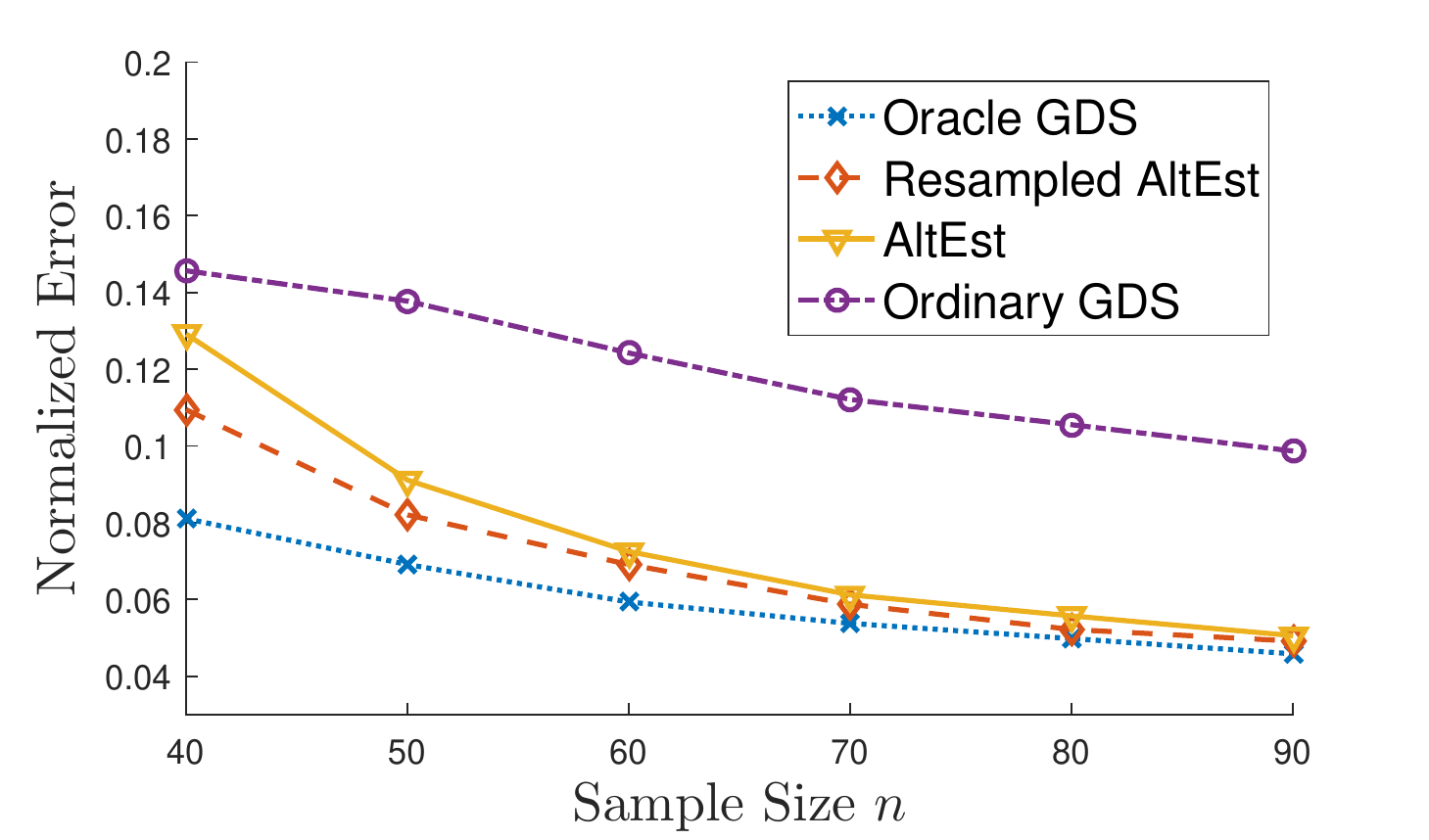}}
\vspace{-3mm}
\caption{(a) When $n = 40$, AltEst is not quite stable due to the large initial error and poor quality of estimated covariance. Then the errors start to decrease for $n \geq 50$.  (b) Resampld AltEst does benefit from fresh samples, and its error is slightly smaller than AltEst as well as more stable when $n$ is small. (c) Oracle GDS outperforms the others, but the performance of AltEst is also competitive. Ordinary GDS is unable to utilize the noise correlation, thus resulting in relatively large error. By comparing the two implementations of AltEst, we can see that resampled AltEst yields smaller error especially when data is inadequate, but their errors are very close if $n$ is suitably large.     \vspace{-3mm}}
\label{fig:exp1}
\end{figure*}

In this section, we present some experimental results to support our theoretical analysis. Specifically we focus on the sparse structure of $\Th^*$ captured by $L_1$ norm. Throughout out the experiment, we fix problem dimension $p=500$,  sparsity level of $\Th^*$ $s = 20$,  and number of iterations for AltEst $T = 5$. Entries of design $\bX$ is generated by i.i.d. standard Gaussians, and $\Th^* = [\underbrace{1, \ldots, 1}_{10}, \underbrace{-1, \ldots, -1}_{10}, \underbrace{0, \ldots, 0}_{480}]^T$. $\SSigma_*$ is
a block diagonal matrix with blocks $\small \SSigma' = \left[ \begin{array}{cc} 1 & 0.8 \\ 0.8 & 1 \end{array}\right]$  duplicated along diagonal, and number of responses $m$ is assumed to be even. All plots are obtained by averaging 100 trials. In the first set of experiments, we set $m = 10$ and investigate the estimation error of $\hat{\Th}_t$ as sample size $n$ varies from 40 to 90. We run AltEst (with and without resampling), the oracle GDS, and the ordinary GDS with $\SSigma = \bI_{m \times m}$. The results are given in Figure \ref{fig:exp1}.

\begin{figure*}[!h]
\centering
\vspace{-3mm}
\subfigure[Error for AltEst] {
    \label{fig2:alt_est_rsp}
    \includegraphics[width=0.35\linewidth]{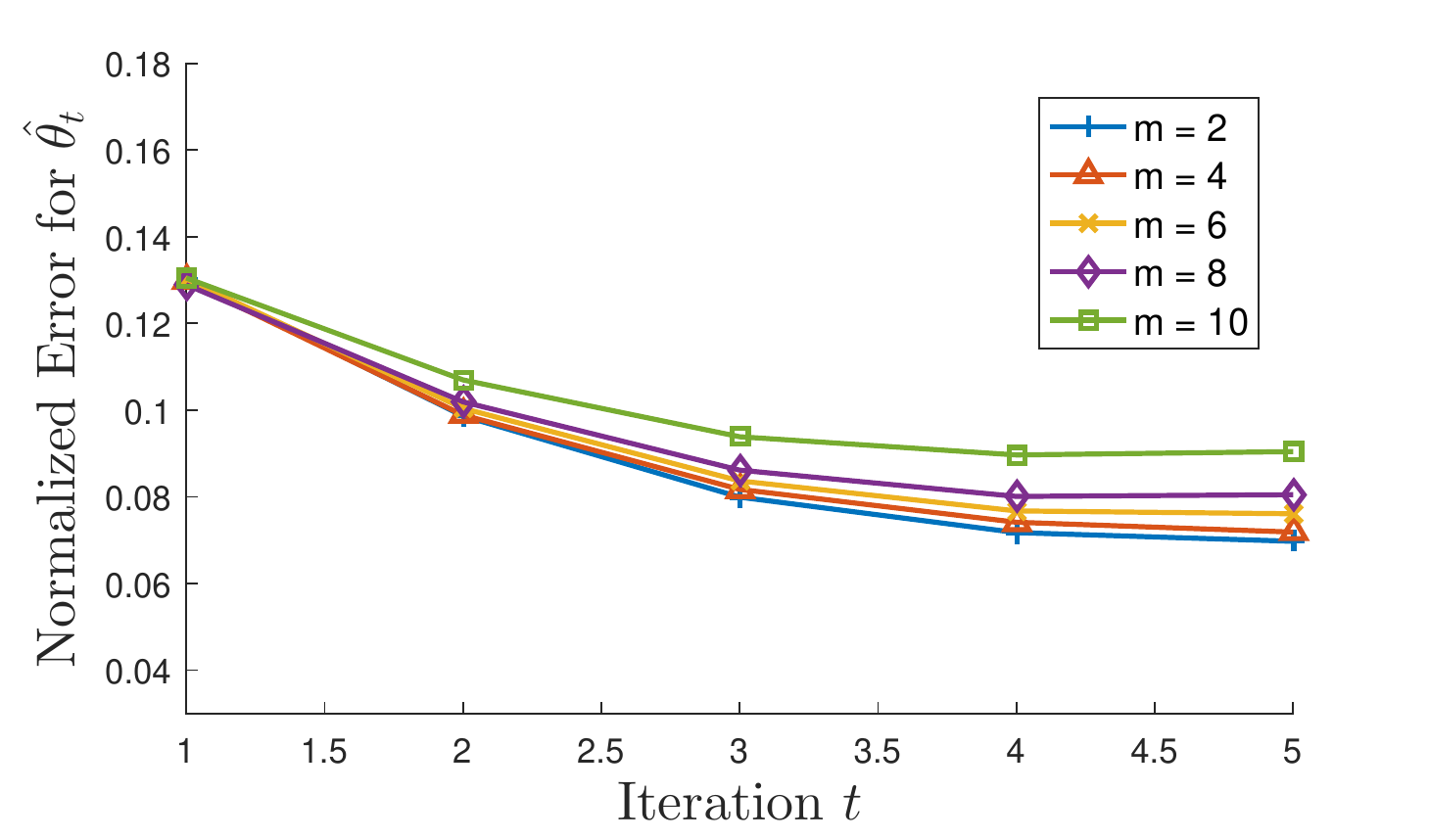}}
\begin{minipage}[b]{0.4in}

\end{minipage}
\subfigure[Comparison of Estimators] {
    \label{fig2:err_cmpr}
    \includegraphics[width=0.35\linewidth]{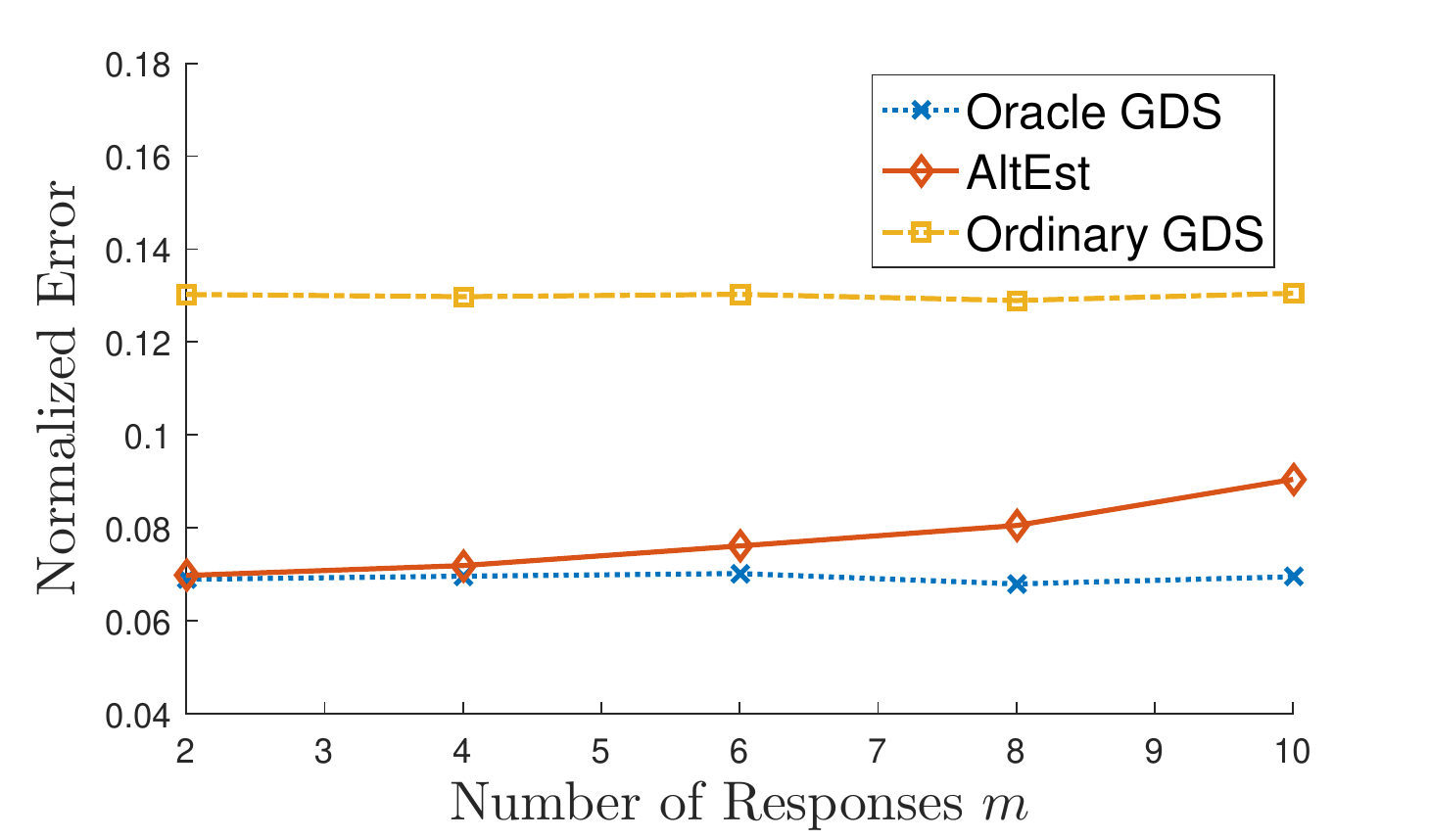}}
\vspace{-3mm}
\caption{(a) Larger error comes with bigger $m$, which confirms that $e_{\min}$ is increasing along with $m$.  (b) The plot for oracle GDS implies that $e_{\text{orc}}$ keeps unchanged, which matches our theoretical result. Though $e_{\min}$ increases, AltEst still outperform the ordinary GDS by a large margin.   \vspace{-3mm}}
\label{fig:exp2}
\end{figure*}

For the second experiment, we fix the product $mn \approx 500$, and let $m = 2, 4, \ldots, 10$. For our choice of $\SSigma_*$, the error incurred by oracle GDS $e_\text{orc}$ is the same for every $m$. We compare AltEst with both oracle and ordinary GDS, and the result is shown in Figure \ref{fig:exp2}.

\newpage

{\LARGE \textbf{Supplementary Material}} 

\setcounter{section}{0}
\renewcommand\thesection{\Alph{section}}
\renewcommand\thesubsection{\thesection.\arabic{subsection}}
\renewcommand*{\thetheorem}{\Alph{theorem}}
\renewcommand{\thelemma}{\Alph{lemma}}
\renewcommand{\theproposition}{\Alph{proposition}}

\section{Preliminaries}
\label{sec:pre}
In this section, we provide some background knowledge and lemmas, which is needed in our proofs. For the sake of convenience, $C$, $C_0$, $c$, $c_0$ and so on are reserved for absolute constants.

\subsection{Sub-Gaussian Random Variable/Vector}
A random variable $x$ is sub-Gaussian if the $\psi_2$-norm defined below is finite
\beq
\vertiii{x}_{\psi_2} = \sup_{q\geq1} \ \frac{\E |x|^q}{\sqrt{q}} < + \infty
\eeq
A random vector $\x \in \R^p$ is sub-Gaussian if $\langle \x, \u \rangle$ is sub-Gaussian for any $\u \in \R^p$, and $\vertiii{\x}_{\psi_2} = \sup_{\u \in \R^p} \vertiii{\langle \x, \u \rangle}_{\psi_2}$. A complete introduction can be found in \cite{vers12}. Here we list some of the well-known properties of sub-Gaussian random variables/vectors, which are extracted from \cite{vers12}.
\begin{proposition}[Sub-Gaussian Tail]
\label{prop:subgauss_tail}
A random variable $x$ satisfies the following inequality iff $\vertiii{x}_{\psi_2} \leq \kappa$,
\beq
\P \left( |x| > \epsilon \right) \leq e \cdot \exp\left(-\frac{C \epsilon^2}{\kappa^2}\right)~,
\eeq
where $C$ is a absolute constant.
\end{proposition}
\begin{proposition}
\label{prop:rot_invar}
If $x_1, x_2, \ldots, x_n$ are independent centered sub-Gaussian random variables, then $\sum_i x_i$ is also a centered sub-Gaussian random variable with
\beq
\vertiii{\sum_{i=1}^n x_i}_{\psi_2}^2 \leq C^2 \sum_{i=1}^n \vertiii{x_i}_{\psi_2}^2 ~,
\eeq
where $C$ is an absolute constant.
\end{proposition}
\begin{proposition}
\label{prop:subgauss_prod}
If $x_1, x_2, \ldots, x_n$ are independent centered sub-Gaussian random variables (not necessarily identical), then $\x = [x_1, \ldots, x_n]^T$ is a centered sub-Gaussian random vector with
\beq
\vertiii{\x}_{\psi_2} \leq C \max_{1 \leq i \leq n} \vertiii{x_i}_{\psi_2} ~,
\eeq
where $C$ is an absolute constant.
\end{proposition}
Essentially Proposition \ref{prop:subgauss_prod} can be shown using the definition of sub-Gaussian vector and Proposition \ref{prop:rot_invar}, which we generalize to independent sub-Gaussian vectors as follows.
\begin{lemma}
\label{lem:subgauss_vec_prod}
If $\x_1, \x_2, \ldots, \x_n$ are all $m$-dimensional independent centered sub-Gaussian random vectors, then $\x = [\x_1^T, \ldots, \x_n^T]^T \in \R^{mn}$ is also a centered sub-Gaussian random vector with
\beq
\vertiii{\x}_{\psi_2} \leq C \max_{1 \leq i \leq n} \vertiii{\x_i}_{\psi_2} ~,
\eeq
where $C$ is an absolute constant.
\end{lemma}
\proof Define $\a = [\a_1^T, \a_2^T, \ldots, \a_n^T]^T \in \s^{mn-1}$, where each $\a_i$ is $m$-dimensional. We have
\begin{align*}
\vertiii{\left\langle \x , \a \right\rangle}_{\psi_2} &= \vertiii{\sum_{i=1}^n \left\langle \x_i , \a_i \right\rangle}_{\psi_2} \leq \sqrt{C^2 \sum_{i=1}^n \vertiii{\left\langle \x_i , \a_i \right\rangle}_{\psi_2}^2} \leq  \sqrt{C^2 \sum_{i=1}^n \|\a_i\|_2^2 \vertiii{ \x_i }_{\psi_2}^2} \\
&\leq
\sqrt{C^2 \sum_{i=1}^n \|\a_i\|_2^2}
\cdot \max_{1\leq i \leq n} \vertiii{ \x_i }_{\psi_2} = C \max_{1 \leq i \leq n} \vertiii{ \x_i }_{\psi_2} ~,
\end{align*}
where we use Proposition \ref{prop:rot_invar} for the first inequality. Based on the definition of sub-Gaussian random vector, we complete the proof. \qed

\subsection{Generic Chaining and Gaussian Width}
One important tool that we use in our probabilistic argument is \emph{generic chaining} \cite{tala05,tala14}, which is powerful for bounding the suprema of stochastic processes. Suppose $\{Z_{\t}\}_{\t \in \mathcal{T}}$ is a centered stochastic process, where each $Z_t$ is a centered random variable. We assume the index set $\mathcal{T}$ is endowed with some metric (distance function) $s(\cdot, \cdot)$. A key notion in generic chaining is $\gamma_2$-functional $\gamma_2(\cT, s)$, which is defined for the metric space $(\cT, s)$. One can think of $\gamma_2$-functional as a measure of the size of set $\cT$ w.r.t. metric $s$. For self-containedness, we give the expression of $\gamma_2(\cT,s)$.
\beq
\gamma_2(\cT,s) = \inf_{\{\cP_n\}} \ \sup_{\t \in \cT} \sum_{n\geq 0} 2^{n / 2} \cdot \diam(\cP_n(\t), s) ~,
\eeq
where $\{\cP_n\}_{n=0}^{\infty} = \{\cP_0, \cP_1, \ldots, \cP_n, \ldots\}$ is a sequence of partitions for $\cT$, which satisfy that $|\cP_0| = 1$, $|\cP_n| \leq 2^{2^n}$ for $n \geq 1$, and that $\cP_{n+1}$ is a finer partition than $\cP_n$, i.e., every $\cQ \in \cP_{n+1}$ is a subset of some $\cQ' \in \cP_n$. $\cP_n(\t)$ denotes the subset of $\cT$ that contains $\t$ in the $n$-th partition, and $\diam(\cP_n(\t), s)$ measures the diameter of $\cP_n(\t)$ w.r.t. metric $s(\cdot, \cdot)$. Note that $\gamma_2$-functional is a purely geometric concept, which involves no probability. Given that $\gamma_2$-functional is fairly involved, we are not going to discuss any insights behind this definition, and refer interested readers to the introductory books \cite{tala05,tala14}. Based on its definition, we list a few straightforward properties of $\gamma_2$-functional here.
\beq
\label{eq:gamma2_1}
\gamma_2(\mathcal{T},s_1) \leq \gamma_2(\mathcal{T}, s_2) \ \ \ \text{if} \ s_1(\u, \v) \leq s_2(\u, \v), \forall \ \u, \v \in \mathcal{T}
\eeq
\beq
\label{eq:gamma2_2}
\gamma_2(\mathcal{T},\beta s) =  \beta \cdot \gamma_2(\mathcal{T},s) \ \ \ \text{for any $\beta > 0$} ~.
\eeq
\beq
\label{eq:gamma2_3}
\gamma_2(\mathcal{T}_1, s_1) = \gamma_2(\mathcal{T}_2, s_2) \ \ \ \text{if} \ \exists \ \text{a global isometry between $(\cT_1, s_1)$ and $(\cT_2, s_2)$}
\eeq
The following lemma concerned with the suprema of $\{Z_{\t}\}$ combines Theorem 2.2.22 and 2.2.27 from \cite{tala14}.
\begin{lemma}
\label{lem:gc}
Given metric space $(\cT, s)$, if the associated centered stochastic process $\{Z_{\t}\}_{\t \in \cT}$ satisfies the condition
\beq
\P \left( |Z_{\u} - Z_{\v} | \geq \epsilon \right) \ \leq \ C_0 \exp \left( - \frac{C_1 \epsilon^2}{s^2(\u, \v)}\right) , \ \ \forall \ \u, \v \in \cT ~,
\eeq
then the following inequalities hold
\beq
\label{eq:gc_bound}
\E \left[ \sup_{\t \in \mathcal{T}} Z_{\t} \right] \leq C_2 \gamma_2 \left(\mathcal{T}, s\right) ~,
\eeq
\beq
\label{eq:gc_conc}
\begin{gathered}
\P \left(\sup_{\u,\v \in \mathcal{T}} |Z_{\u} - Z_{\v} | \geq C_3 \left(\gamma_2(\mathcal{T}, s) + \epsilon \cdot \diam(\mathcal{T},s) \right) \right) \leq C_4 \exp \left( -\epsilon^2 \right) ~,
\end{gathered}
\eeq
where $C_0, C_1, C_2, C_3$ and $C_4$ are all absolute constants.
\end{lemma}
Another useful result based on generic chaining is the Theorem D in \cite{mept07}.
\begin{lemma}[Theorem D in \cite{mept07}]
\label{lem:mept}
There exist absolute constants $C_1$, $C_2$ for which the following
holds. Let $(\Omega,\mu)$ be a probability space on which $X$ is defined, and $X_1, \ldots, X_n$ be independent copies of $X$. Let set $\cH$ be a subset of the unit
sphere of $L_2(\mu)$, i.e., $\cH \subseteq \s_{L_2} = \{ h : \vertiii{h}_{L_2} = \sqrt{\int_{\Omega} h^2(X) dX} = 1\}$, and assume that $ \sup_{h \in \cH}~\vertiii{h}_{\psi_2} \leq \kappa$. Then, for any $\beta> 0$ and $n \geq 1$ satisfying
\beq
\label{eq:beta_cond}
C_1 \kappa \gamma_2(\cH, \vertiii{\cdot}_{\psi_2}) \leq \beta \sqrt{n}~,
\eeq
with probability at least $1- \exp(-C_2 \beta^2 n/\kappa^4)$,
\beq
\sup_{h \in \cH}~\left| \frac{1}{n} \sum_{i=1}^n h^2(X_i) - \E\left[h^2\right] \right| \leq \beta~.
\eeq
\end{lemma}
The suprema in both Lemma \ref{lem:gc} and \ref{lem:mept} are characterized in terms of $\gamma_2$-functional, which is not easily computable. In order to further bound the $\gamma_2$-functional, one needs the so-called \emph{majorizing measures theorem} \cite{tala92}.
\begin{lemma}
\label{lem:mm}
Given any Gaussian process $\{Y_{\t}\}_{\t \in \mathcal{T}}$, define $s(\u,\v) = \sqrt{\E |Y_{\u} - Y_{\v} |^2}$ for $\u, \v \in \mathcal{T}$. Then $\gamma_2(\mathcal{T}, s)$ can be upper bounded by
\beq
\gamma_2(\mathcal{T},s) \leq C_0 \E \left[ \sup_{\t \in \mathcal{T}} Y_{\t} \right]  ~,
\eeq
where $C_0$ is an absolute constant.
\end{lemma}
We construct the simple Gaussian process $\{Y_{\t} = \langle \t, \g \rangle \}_{\t \in \cT}$ for any $\cT \subseteq \R^p$, where $\g$ is a standard Gaussian random vector. Hence $s(\u,\v) = \sqrt{\E |Y_{\u} - Y_{\v} |^2} = \sqrt{\E |\langle \u -\v, \g \rangle |^2} = \|\u - \v\|_2$. It follows from Lemma \ref{lem:mm} that
\beq
\label{eq:gamma2_width}
\gamma_2 \left(\cT, \|\cdot\|_2\right) \leq C_0 \E \left[ \sup_{\t \in \cT} \ \langle \t, \g \rangle \right] = C_0 \cdot w(\cT) ~,
\eeq
which makes the connection between $\gamma_2$-functional and Gaussian width. One technique we utilize in our proof for bounding Gaussian width is as follows, which originates in \cite{mapr14}.
\begin{lemma}[Lemma 2 in \cite{mapr14}]
\label{lem:union_width}
Let $M > 4$, $\cA_1, \cdots, \cA_M \subset \R^p$, and $\cA = \cup_m \cA_m$. The Gaussian width of  $\cA$ satisfies
\beq
w(\cA) \leq \max_{1 \leq m \leq M} w(\cA_m) + 2 \sup_{\z \in \cA} \|\z\|_2 \sqrt{\log M}
\eeq
\end{lemma}

\subsection{Proof of Lemma \ref{lem:subgauss_exp}}

\proof Let $\w = \XXi^{\frac{1}{2}} \u$ for any $\u \in \s^{m-1}$, and we have
\begin{align*}
\GGamma_{\u} &= \E\left[\LLambda^{\frac{1}{2}} \tilde{\bX}^T \XXi^{\frac{1}{2}} \u \u^T \XXi^{\frac{1}{2}} \tilde{\bX} \LLambda^{\frac{1}{2}}\right] \\ &=
\E\left[ \left[\LLambda^{\frac{1}{2}} \tilde{\x}_1, \ldots, \LLambda^{\frac{1}{2}} \tilde{\x}_m\right] \cdot \left[ \begin{array}{ccc} w_1 \\ \vdots \\ w_m \end{array} \right] \cdot \left[w_1, \ldots, w_m\right] \cdot \left[ \begin{array}{ccc} \tilde{\x}_1^T \LLambda^{\frac{1}{2}} \\ \vdots \\ \tilde{\x}_m^T \LLambda^{\frac{1}{2}} \end{array} \right] \right] \\
&= \sum_{i=1}^m \sum_{j=1}^m w_i w_j \E\left[ \LLambda^{\frac{1}{2}}\tilde{\x}_i \tilde{\x}_j^T \LLambda^{\frac{1}{2}} \right] = \sum_{i=1}^m w_i^2 \LLambda^{\frac{1}{2}} \E\left[ \tilde{\x}_i \tilde{\x}_i^T \right] \LLambda^{\frac{1}{2}} = \left\| \XXi^{\frac{1}{2}} \u \right\|_2^2 \cdot \LLambda
\end{align*}
It is clear that
\begin{align*}
\lambda_{\min}(\XXi) \cdot \lambda_{\min}(\LLambda) \leq \lambda_{\min}(\GGamma_{\u}) \leq \lambda_{\max}(\GGamma_{\u}) \leq \lambda_{\max}(\XXi) \cdot \lambda_{\max}(\LLambda) ~,
\end{align*}
which indicates that condition \eqref{eq:mu_max_min} holds.
If $\vertiii{\tilde{\x}_i}_{\psi_2} \leq \tilde{\kappa}$, then
\begin{align*}
\vertiii{\bX}_{\psi_2} &= \sup_{\substack{\v \in \s^{p-1} \\ \u \in \s^{m-1}}} \vertiii{\v^T \GGamma_{\u}^{-\frac{1}{2}} \bX^T \u}_{\psi_2} = \sup_{\substack{\v \in \s^{p-1} \\ \u \in \s^{m-1}}} \vertiii{\frac{\v^T \LLambda^{-\frac{1}{2}}}{\| \XXi^{\frac{1}{2}} \u\|_2} \cdot \LLambda^{\frac{1}{2}} \tilde{\bX}^T \XXi^{\frac{1}{2}}  \u}_{\psi_2} \\
&= \sup_{\substack{\v \in \s^{p-1} \\ \u \in \s^{m-1}}} \vertiii{\frac{\v^T \tilde{\bX}^T}{\| \XXi^{\frac{1}{2}} \u\|_2} \cdot  \XXi^{\frac{1}{2}}  \u}_{\psi_2} = \sup_{\substack{\v \in \s^{p-1}}}  \vertiii{\tilde{\bX} \v}_{\psi_2} \leq C \tilde{\kappa}
\end{align*}
where the inequality follows from noting that the vector $\tilde{\bX} \v$ has independent elements with $\psi_2$-norm bounded by $\tilde{\kappa}$, and thus $\vertiii{\tilde{\bX} \v}_{\psi_2} \leq C \tilde{\kappa}$ for any $\v \in \s^{p-1}$. Therefore condition \eqref{eq:subgauss_mat} also holds with $\kappa = C \tilde{\kappa}$. \qed

\section{Proofs for Section \ref{sec:gds_analysis}}
\subsection{Proof of Lemma \ref{lem:gds}}

\proof Since $\hat{\Th}$ is feasible and $\gamma_n$ is selected to be admissible,
we have
\begin{gather*}
\left\| \frac{1}{n} \sum_{i=1}^n \bX^T_i \SSigma^{-1} (\bX_i \hat{\Th} - \y_i) \right\|_* \leq \gamma_n, \quad \left\| \frac{1}{n} \sum_{i=1}^n \bX^T_i \SSigma^{-1} (\bX_i \Th^* - \y_i) \right\|_* \leq \gamma_n \\
\Longrightarrow \quad \left\| \frac{1}{n} \sum_{i=1}^n \bX^T_i \SSigma^{-1} \bX_i (\hat{\Th} - \Th^*) \right\|_* \leq 2 \gamma_n \\
\Longrightarrow \quad \left\langle \hat{\Th} - \Th^*, \frac{1}{n} \sum_{i=1}^n \bX^T_i \SSigma^{-1} \bX_i (\hat{\Th} - \Th^*) \right\rangle \leq \|\hat{\Th} - \Th^*\| \cdot \left\| \frac{1}{n} \sum_{i=1}^n \bX^T_i \SSigma^{-1} \bX_i (\hat{\Th} - \Th^*) \right\|_* \\
\Longrightarrow \quad (\hat{\Th} - \Th^*)^T \left( \frac{1}{n} \sum_{i=1}^n \bX^T_i \SSigma^{-1} \bX_i \right) (\hat{\Th} - \Th^*) \leq 2 \gamma_n \|\hat{\Th} - \Th^*\|
\end{gather*}
As $\|\hat{\Th}\| \leq \|\Th^*\|$, we have $\frac{\hat{\Th} - \Th^*}{\|\hat{\Th} - \Th^*\|_2} \in \cA\left(\Th^*\right)$. By the assumption of RE condition, we further obtain
\begin{gather*}
\alpha \| \hat{\Th} - \Th^* \|_2^2 \leq (\hat{\Th} - \Th^*)^T \left( \frac{1}{n} \sum_{i=1}^n \bX^T_i \SSigma^{-1} \bX_i \right) (\hat{\Th} - \Th^*) \leq 2 \gamma_n \|\hat{\Th} - \Th^*\| \\
\Longrightarrow \quad \| \hat{\Th} - \Th^* \|_2 \leq \frac{\| \hat{\Th} - \Th^* \|}{\| \hat{\Th} - \Th^* \|_2} \cdot \frac{2\gamma_n}{\alpha} \leq 2 \Psi(\Th^*) \cdot \frac{\gamma_n}{\alpha} ~,
\end{gather*}
where we use the definition of restricted norm compatibility. \qed

\subsection{Proof of Lemma \ref{lem:re}}

\proof Assume that the eigenvalue decomposition of $\SSigma$ is given by $\SSigma = \sum_{i=j}^m \sigma_i \u_j \u_j^T$. For convenience, we denote $\z^j = \bX^T \u_j$, $\z^j_i = \bX_i^T \u_j$, and $\hat{\GGamma}_j = \frac{1}{n} \sum_{i=1}^n \bX_i^T \u_j \u_j^T \bX_i$. Note that $\GGamma_j = \E [\z^j \z^{j^T}]$, $\GGamma = \sum_{i=j}^m \frac{\GGamma_j}{\sigma_j}$, $\hat{\GGamma}_j = \frac{1}{n} \sum_{i=1}^n \z_i^j \z_i^{j^T}$, and $\hat{\GGamma} =  \sum_{j=1}^m \frac{\hat{\GGamma_j}}{\sigma_j}$. In order to apply Lemma \ref{lem:mept}, we let $(\Omega_j, \mu_j)$ be the probability measure that $\z^j$ is defined on, and construct the function set
\begin{align*}
\cH_j = \left\{ h_{\v}  = \left\langle \GGamma_j^{-\frac{1}{2}} \v, \cdot \right\rangle \ | \ \v \in \cA_{\GGamma_j} \right\}
\end{align*}
It is easy to see that for any $h_{\v} \in \cH_j$,
\begin{align*}
\E [h_{\v}^2] = \E_{\z^j \sim \mu_j} \left[\v^T \GGamma_j^{-\frac{1}{2}} \z^j \z^{j^T} \GGamma_j^{-\frac{1}{2}} \v \right] = \v^T \GGamma_j^{-\frac{1}{2}} \left( \E_{\z^j \sim \mu_j} \left[ \z^j \z^{j^T} \right] \right) \GGamma_j^{-\frac{1}{2}} \v = \v^T \v = 1 ~,
\end{align*}
i.e., $\cH_j \subseteq \s_{L_2(\mu_j)} = \{ h \ | \ \vertiii{h}_{L_2(\mu_j)} = 1\}$. Based on the definition of sub-Gaussian $\bX$, we also have for any $\v \in \cA_{\GGamma_j}$,
\begin{align*}
\vertiii{h_{\v}}_{\psi_2} = \vertiii{\left\langle \GGamma^{-\frac{1}{2}}_j \v, \z^j \right\rangle}_{\psi_2} = \vertiii{\v^T \GGamma^{-\frac{1}{2}}_j \bX^T \u_j}_{\psi_2} \leq \kappa ~,
\end{align*}
and also for any $\v_1, \v_2 \in \cA_{\GGamma_j}$, we have
\begin{align*}
\vertiii{h_{\v_1} - h_{\v_2}}_{\psi_2} = \vertiii{(\v_1 - \v_2)^T \GGamma^{-\frac{1}{2}}_j \z^j}_{\psi_2} \leq \kappa \cdot \|\v_1 - \v_2\|_2 ~.
\end{align*}
If we choose $\beta = \frac{1}{2}$, using \eqref{eq:gamma2_1}, \eqref{eq:gamma2_2} and \eqref{eq:gamma2_3}, then we have
\begin{align*}
c_1 \kappa \cdot \gamma_2(\cH_j, \vertiii{\cdot}_{\psi_2}) \leq c_1 \kappa^2 \cdot \gamma_2(\cA_{\GGamma_j}, \|\cdot\|_2) \leq c_1 c_4 \kappa^2 \cdot w(\cA_{\GGamma_j}) \leq \beta \sqrt{n}
\end{align*}
when $n \geq C_1 \kappa^4 w^2(\cA_{\GGamma_j})$ where $C_1 = 4 c_1^2 c_4^2$. By Lemma \ref{lem:mept}, with probability at least $1 - \exp(-c_2 \beta^2 n / \kappa^4) = 1 - \exp(-C_2 n / \kappa^4)$ where $C_2 = c_2 / 4$, we have
\begin{align*}
\sup_{h \in \cH_j} \left|\frac{1}{n} \sum_{i=1}^n h^2(\z^j_i) - \E[h^2] \right| &=  \sup_{\v \in \cA_{\GGamma_j}} \left|\frac{1}{n} \sum_{i=1}^n \v^T \GGamma^{-\frac{1}{2}}_j \z^j_i \z^{j^T}_i \GGamma^{-\frac{1}{2}}_j \v - 1 \right| \\
&= \sup_{\v \in \cA_{\GGamma_j}} \left|\v^T \GGamma^{-\frac{1}{2}}_j \hat{\GGamma}_j \GGamma^{-\frac{1}{2}}_j \v - 1 \right| \leq \frac{1}{2}
\end{align*}
\vspace{-3mm}
\begin{gather*}
\Longrightarrow \quad  \v^T \GGamma^{-\frac{1}{2}}_j \hat{\GGamma}_j \GGamma^{-\frac{1}{2}}_j \v  \geq \frac{1}{2}, \quad \forall \ \v \in \cA_{\GGamma_j} \\
\Longrightarrow 
\v^T \GGamma^{-\frac{1}{2}}_j \hat{\GGamma}_j \GGamma^{-\frac{1}{2}}_j \v  \geq \frac{1}{2} \left(\v^T \GGamma^{-\frac{1}{2}}_j \GGamma_j \GGamma^{-\frac{1}{2}}_j \v \right), \quad \forall \ \v \in \cA_{\GGamma_j}
\end{gather*}
Let $\w =  \GGamma^{-\frac{1}{2}}_j \v$, and note that the inequalities above are preserved under arbitrary scaling of $\w$. By recalling the definition of $\cA_{\GGamma_j}$, it is not difficult to see that
\beq
\label{eq:re_uu}
\w^T \hat{\GGamma}_j \w \geq \frac{1}{2} \w^T \GGamma_j \w , \quad \forall \ \w \in \cA ~.
\eeq
Combining \eqref{eq:re_uu} for each $\GGamma_j$ using union bound, we obtain
\begin{gather*}
\w^T \left(\sum_{i=1}^m \frac{\hat{\GGamma}_j}{\sigma_j}\right) \w \geq \frac{1}{2} \w^T \left(\sum_{i=1}^m \frac{\GGamma_j}{\sigma_j}\right) \w , \ \ \forall \ \w \in \cA \quad
\Longrightarrow \quad
\w^T \hat{\GGamma} \w \geq \frac{1}{2} \w^T \GGamma \w , \ \ \forall \ \w \in \cA ~,
\end{gather*}
which completes the proof by renaming $\w$ as $\v$. \qed

\subsection{Proof of Lemma \ref{lem:width_rel}}
\proof Recall the definition of Gaussian width $w(\cA_{\GGamma_{\u}}) = \E\left[\sup_{\v \in \cA_{\GGamma_{\u}}} \langle \v, \g \rangle\right]$, where $\g$ is a standard Gaussian random vector. Given the assumption \eqref{eq:mu_max_min}, we have $\mu_{\min} \leq \lambda_{\min}(\GGamma_{\u}) \leq \lambda_{\max} (\GGamma_{\u}) \leq \mu_{\max}$, and note that
\begin{align}
\label{eq:width_rel_1}
\begin{split}
\sup_{\v \in \cA_{\GGamma_{\u}}} \langle \v, \g  \rangle &= \sup_{\v \in \cA_{\GGamma_{\u}}} \left\langle \GGamma_{\u}^{-\frac{1}{2}}\v, \GGamma_{\u}^{\frac{1}{2}} \g  \right\rangle
\leq \sup_{\v \in \cone(\cA) \cap \frac{1}{\sqrt{\mu_{\min}}}\B^p} \left\langle \v, \GGamma_{\u}^{\frac{1}{2}} \g \right \rangle \\
&=
\frac{1}{\sqrt{\mu_{\min}}} \cdot \sup_{\v \in \cone(\cA) \cap \B^p} \left\langle \v, \GGamma_{\u}^{\frac{1}{2}} \g \right \rangle ~,
\end{split}
\end{align}
where the inequality follows from $\GGamma_{\u}^{-\frac{1}{2}}\v \in \cone(\cA)$ and $\|\GGamma_{\u}^{-\frac{1}{2}}\v\|_2 \leq \frac{1}{\sqrt{\mu_{\min}}}$. Now we use generic chaining to bound the right-hand side above. Denote the set $\cone(\cA) \cap \B^p$ by $\cT$, and we consider the stochastic process $\{Z_{\v} = \langle \v, \GGamma_{\u}^{\frac{1}{2}} \g \rangle \}_{\v \in \cT}$. For any $\v_1, \v_2 \in \cT$, we have
\begin{align*}
\vertiii{Z_{\v_1} - Z_{\v_2}}_{\psi_2} = \vertiii{\langle \GGamma_{\u}^{\frac{1}{2}} (\v_1 - \v_2), \g \rangle}_{\psi_2} \leq \kappa_0 \left\|\GGamma_{\u}^{\frac{1}{2}} (\v_1 - \v_2) \right\|_2  \leq \kappa_0  \sqrt{\mu_{\max}} \cdot \|\v_1 - \v_2\|_2 ~.
\end{align*}
If we define for $\cT$ the metric $s(\v_1, \v_2) = \kappa_0  \sqrt{\mu_{\max}} \cdot \|\v_1 - \v_2\|_2$, it follows from Proposition \ref{prop:subgauss_tail} that
\begin{align*}
\P \left( \left| Z_{\v_1} - Z_{\v_2} \right| \geq \epsilon \right) \leq e \cdot \exp\left(- \frac{c \epsilon^2}{\kappa_0^2 \mu_{\max} \|\v_1-\v_2\|_2^2}\right) = e \cdot \exp\left(- \frac{c \epsilon^2}{s^2(\v_1, \v_2)}\right) ~.
\end{align*}
By Lemma \ref{lem:gc}, \eqref{eq:gamma2_2} and \eqref{eq:gamma2_width}, we obtain
\begin{align}
\label{eq:width_rel_2}
\begin{split}
\E\left[\sup_{\v \in \cT} \langle \v, \GGamma_{\u}^{\frac{1}{2}} \g \rangle \right] &= \E\left[\sup_{\v \in \cT} Z_{\v}\right] \leq c_1 \gamma_2(\cT, s) = c_1 \kappa_0 \sqrt{\mu_{\max}}\gamma_2(\cT, \|\cdot\|_2) \leq c_1 c_2 \kappa_0 \sqrt{\mu_{\max}} \cdot w(\cT)
\end{split}
\end{align}
Note that $\cT = \cone(\cA) \cap \B^p \subseteq \conv(\cA \cup \{\mathbf{0}\})$. By Lemma \ref{lem:union_width}, we have
\begin{align}
\label{eq:width_rel_3}
w(\cT) \leq w(\conv(\cA \cup \{\mathbf{0}\})) = w(\cA \cup \{\mathbf{0}\}) \leq \max\left\{w(\cA), w(\mathbf{0}) \right\}  + 2 \sqrt{\ln 4} \leq w(\cA) + 3 ~.
\end{align}
Combining \eqref{eq:width_rel_1}, \eqref{eq:width_rel_2} and \eqref{eq:width_rel_3}, we have
\begin{align}
w(\cA_{\GGamma_{\u}}) = \E \left[\sup_{\v \in \cA_{\GGamma_{\u}}} \langle \v, \g  \rangle\right] \leq \frac{1}{\sqrt{\mu_{\min}}} \E \left[ \sup_{\v \in \cT} \left\langle \v, \GGamma_{\u}^{\frac{1}{2}} \g \right\rangle \right] \leq c_1 c_2 \kappa_0 \sqrt{\frac{\mu_{\max}}{\mu_{\min}}} \cdot \left(w(\cA) + 3\right) ~,
\end{align}
where the last inequality follows from condition \eqref{eq:mu_max_min}. \qed

\subsection{Proof of Corollary \ref{cor:re}}

\proof Given the definition of sub-Gaussian $\bX$ and Lemma \ref{lem:re}, we have
\begin{align*}
\v^T \hat{\GGamma} \v &\geq \frac{1}{2}\v^T \GGamma \v = \frac{1}{2}\v^T \left( \sum_{j=1}^m \frac{1}{\sigma_j} \cdot \E\left[\bX^T \u_j \u^T_j \bX \right] \right) \v \\
&\geq \frac{\mu_{\min}}{2} \cdot \v^T \v \left( \sum_{j=1}^m \frac{1}{\sigma_j} \right) = \frac{\mu_{\min}}{2} \tr\left(\SSigma^{-1}\right) ~.
\end{align*}
Using the bound in Lemma \ref{lem:width_rel}, we have
\begin{align*}
n \geq C_1 \kappa_0^2 \kappa^4 \cdot  \frac{\mu_{\max}}{\mu_{\min}} \cdot (w(\cA) + 3)^2 \quad \Longrightarrow \quad n \geq C \kappa^4 \cdot \max_j \left\{w^2(\cA_{\GGamma_j})\right\}
\end{align*}
We complete the proof by combining the two equations above. \qed

\subsection{Proof of Lemma \ref{lem:gamma_n}}

\proof Since design $\bX_i$ and noise $\Eta_i$ are independent, we first consider the scenario where each $\Eta_i$ is arbitrary but fixed vector. Using the definition of dual norm, we have
\begin{align*}
\left\|\frac{1}{n} \sum_{i=1}^n \bX_i^T \SSigma^{-1} \Eta_i  \right\|_* &= \frac{1}{n} \cdot \sup_{\v \in \cB} \left\langle \v, \ \sum_{i=1}^n \bX_i^T \SSigma^{-1} \Eta_i \right\rangle = \frac{1}{n} \cdot \sup_{\v \in \cB} \sum_{i=1}^n \left\langle \LLambda_i^{\frac{1}{2}} \v, \ \LLambda_i^{-\frac{1}{2}} \bX_i^T \SSigma^{-1} \Eta_i \right\rangle
\end{align*}
where $\LLambda_i = \E_{\bX_i} [\bX^T_i \SSigma^{-1} \Eta_i \Eta_i^T \SSigma^{-1} \bX_i ]$. Based on the definition of sub-Gaussian $\bX_i$, we get
\begin{align*}
& \vertiii{\LLambda_i^{-\frac{1}{2}} \bX_i^T \SSigma^{-1} \Eta_i}_{\psi_2} \leq \kappa \quad \Longrightarrow \quad \\
\vertiii{\sum_{i=1}^n \left\langle \LLambda_i^{\frac{1}{2}} \v, \ \LLambda_i^{-\frac{1}{2}} \bX_i^T \SSigma^{-1} \Eta_i \right\rangle}_{\psi_2} &\leq c_0 \max_{1 \leq i \leq n} \vertiii{\LLambda_i^{-\frac{1}{2}} \bX_i^T \SSigma^{-1} \Eta_i}_{\psi_2} \cdot \sqrt{\sum_{i=1}^n \left\| \LLambda_i^{\frac{1}{2}} \v \right\|_2^2} \\
\leq c_0 \kappa &\sqrt{\sum_{i=1}^n \left\| \LLambda_i^{\frac{1}{2}} \right\|_2^2 \|\v\|_2^2} \ \leq \  c_0 \kappa \sqrt{\mu_{\max}} \cdot \sqrt{\sum_{i=1}^n \left\|\SSigma^{-1} \Eta_i\right\|_2^2} \cdot \|\v\|_2
\end{align*}
where we use Lemma \ref{lem:subgauss_vec_prod} in the first inequality by treating the sum of inner products as one ``big'' inner product. The last inequality follows from the definition of $\mu_{\max}$ in \eqref{eq:mu_max_min}.
Now we consider the stochastic process $\left\{Z_{\v} = \left\langle \v, \ \sum_{i=1}^n \bX_i^T \SSigma^{-1} \Eta_i \right\rangle \right\}_{\v \in \cB}$, where $\Eta_i$ is still fixed. For any $Z_{\v_1}$ and $Z_{\v_2}$, by the argument above and Proposition \ref{prop:subgauss_tail}, we have
\begin{gather*}
\vertiii{Z_{\v_1} - Z_{\v_2}}_{\psi_2} \leq c_0 \kappa \sqrt{\mu_{\max}} \cdot \sqrt{\sum_{i=1}^n \left\|\SSigma^{-1} \Eta_i\right\|_2^2} \cdot \|\v_1 - \v_2\|_2 \triangleq s(\v_1, \v_2)  \\
\Longrightarrow \quad \P \left( \left| Z_{\v_1} - Z_{\v_2} \right| > \epsilon \right) \leq e \cdot \exp \left( -\frac{C_1 \epsilon^2}{s^2(\v_1, \v_2)} \right)
\end{gather*}
It follows from \eqref{eq:gamma2_2}, \eqref{eq:gamma2_width} and Lemma \ref{lem:gc} that
\begin{gather*}
\gamma_2(\cB, s) =  c_0 \kappa \sqrt{\mu_{\max}} \cdot \sqrt{\sum_{i=1}^n \left\|\SSigma^{-1} \Eta_i\right\|_2^2} \cdot \gamma_2(\cB, \|\cdot\|_2) \leq  c_0 c_1 \kappa \sqrt{\mu_{\max}} \cdot \sqrt{\sum_{i=1}^n \left\|\SSigma^{-1} \Eta_i\right\|_2^2} \cdot  w(\cB) ~,\\
\P_{\bX_i} \left( \sup_{\v_1, \v_2 \in \cB} \left| Z_{\v_1} - Z_{\v_2} \right|  \geq c_2 \left( \gamma_2(\cB, s) + \epsilon \cdot \diam (\cB, s) \right) \right) \leq c_3 \exp\left(-\epsilon^2\right)
\end{gather*}
Combining the two inequalities above with the symmetry of $\cB$, we obtain
\begin{align*}
\P_{\bX} \left( \sup_{\v \in \cB} Z_{\v}  \geq  c_0 c_2\kappa \sqrt{\mu_{\max}} \cdot \sqrt{\sum_{i=1}^n \left\|\SSigma^{-1} \Eta_i\right\|_2^2} \left( \frac{c_1}{2} \cdot w(\cB) + \epsilon \cdot  \sup_{\v \in \cB} \|\v\|_2 \right) \right) \leq c_3 \exp\left(-\epsilon^2\right)
\end{align*}
Letting $\rho = \sup_{\v \in \cB} \|\v\|_2$, $\epsilon = \frac{c_1 w(\cB)}{2\rho}$, with probability at least $1 - c_3\exp(-\frac{c_1^2 w^2(\cB)}{4 \rho^2})$, we have
\beq
\label{eq:dantzig_1}
\sup_{\v \in \cB} Z_{\v} = \left\|\sum_{i=1}^n \bX_i^T \SSigma^{-1} \Eta_i  \right\|_* \leq c_0 c_1 c_2 \kappa \sqrt{\mu_{\max}} \cdot \sqrt{\sum_{i=1}^n \left\|\SSigma^{-1} \Eta_i\right\|_2^2}  \cdot w(\cB)
\eeq
for any given set of $\Eta_i$. Now we incorporate the randomness of $\Eta_i$. Essentially we need to bound
\begin{align*}
\sqrt{\sum_{i=1}^n \left\|\SSigma^{-1} \Eta_i\right\|_2^2} = \sqrt{\sum_{i=1}^n \left\|\SSigma^{-1} \SSigma_*^{\frac{1}{2}} \tilde{\Eta}_i\right\|_2^2} ~,
\end{align*}
where each $\tilde{\Eta}_i$ is an $m$-dimensional standard (isotropic) Gaussian random vector. Given $\v = [\v_1^T, \ldots, \v_n^T]^T \in \R^{mn}$, Denote $f(\v) = \sqrt{\sum_{i=1}^n \left\|\SSigma^{-1} \SSigma_*^{\frac{1}{2}} \v_i\right\|_2^2}$, and we have
\begin{align*}
\left| f(\v) - f(\w) \right|  &= \left| \ \sqrt{\sum_{i=1}^n \left\|\SSigma^{-1} \SSigma_*^{\frac{1}{2}} \v_i\right\|_2^2} - \sqrt{\sum_{i=1}^n \left\|\SSigma^{-1} \SSigma_*^{\frac{1}{2}} \w_i\right\|_2^2} \ \right| \\
&\leq \sqrt{\sum_{i=1}^n \left(\left\|\SSigma^{-1} \SSigma_*^{\frac{1}{2}} \v_i\right\|_2 - \left\|\SSigma^{-1} \SSigma_*^{\frac{1}{2}} \w_i\right\|_2\right)^2} \\
&\leq \sqrt{\sum_{i=1}^n \left\|\SSigma^{-1} \SSigma_*^{\frac{1}{2}} (\v_i - \w_i)\right\|_2^2} \\
&\leq \sqrt{\sum_{i=1}^n \left\|\SSigma^{-1} \SSigma_*^{\frac{1}{2}}\right\|_2^2 \left\|\v_i - \w_i \right\|_2^2 } = \left\|\SSigma^{-1} \SSigma_*^{\frac{1}{2}}\right\|_2 \|\v -\w\|_2
\end{align*}
which implies that $f$ is a Lipschitz function with parameter $\|\SSigma^{-1} \SSigma_*^{\frac{1}{2}} \|_2$. The first two inequalities use the triangular inequality for $L_2$ norm. Letting $\tilde{\Eta} = [\tilde{\Eta}_1^T, \ldots, \tilde{\Eta}_n^T]^T$, by the concentration inequality for Lipschitz function of Gaussian random vector (see Proposition 5.34 in \cite{vers12}), we obtain
\begin{gather*}
\P \left(f(\tilde{\Eta}) - \E f(\tilde{\Eta}) > t \right) \leq \exp\left(  \frac{-t^2}{2\|\SSigma^{-1} \SSigma_*^{\frac{1}{2}} \|_2^2} \right) \\
\Longrightarrow \ \P \left( \sqrt{\sum_{i=1}^n \left\|\SSigma^{-1} \SSigma_*^{\frac{1}{2}} \tilde{\Eta}_i\right\|_2^2}  - \E \sqrt{\sum_{i=1}^n \left\|\SSigma^{-1} \SSigma_*^{\frac{1}{2}} \tilde{\Eta}_i\right\|_2^2}   > t\right) \leq \exp\left( \frac{-t^2}{2\|\SSigma^{-1} \SSigma_*^{\frac{1}{2}} \|_2^2} \right) \\
\Longrightarrow \  \P \left( \sqrt{\sum_{i=1}^n \left\|\SSigma^{-1} \Eta_i\right\|_2^2} - \sqrt{\E \sum_{i=1}^n\tr\left(\SSigma^{-1} \SSigma_*^{\frac{1}{2}} \tilde{\Eta}_i  \tilde{\Eta}_i^T \SSigma_*^{\frac{1}{2}} \SSigma^{-1}\right) }  > t\right) \leq \exp\left(  \frac{-t^2}{2\|\SSigma^{-1} \SSigma_*^{\frac{1}{2}} \|_2^2} \right) \\
\Longrightarrow \  \P \left( \sqrt{\sum_{i=1}^n \left\|\SSigma^{-1} \Eta_i\right\|_2^2} - \sqrt{n}\sqrt{ \tr\left(\SSigma^{-1}  \SSigma_* \SSigma^{-1}\right) }  > t\right) \leq \exp\left( \frac{-t^2}{2\|\SSigma^{-1} \SSigma_*^{\frac{1}{2}} \|_2^2} \right)
\end{gather*}
where we use Jensen's inequality in the third step for bounding the expectation $\E f(\tilde{\Eta})$. Letting $t = \sqrt{\tr\left(\SSigma^{-1}  \SSigma_* \SSigma^{-1}\right) \cdot  n}$ and $\tau = \|\SSigma^{-1} \SSigma_*^{\frac{1}{2}}\|_{F} / \|\SSigma^{-1} \SSigma_*^{\frac{1}{2}}\|_{2}$, with probability at least $1 - \exp \left(-\frac{n \tau^2}{2} \right)$, we have
\beq
\label{eq:dantzig_2}
\sqrt{\sum_{i=1}^n \left\|\SSigma^{-1} \Eta_i\right\|_2^2} \leq 2\sqrt{n} \cdot \sqrt{ \tr\left(\SSigma^{-1}  \SSigma_* \SSigma^{-1}\right) } ~,
\eeq
where we use the relation $\tr\left(\SSigma^{-1}  \SSigma_* \SSigma^{-1}\right) = \|\SSigma^{-1} \SSigma_*^{\frac{1}{2}}\|_{F}^2$. By applying a union bound to \eqref{eq:dantzig_1} and \eqref{eq:dantzig_2}, with probability at least $1 - \exp\left(-\frac{n \tau^2}{2}\right) - c_3\exp(-\frac{c_1^2 w^2(\cB)}{4 \rho^2})$, the following inequality holds
\beq
\label{eq:dantzig_3}
\left\|\frac{1}{n} \sum_{i=1}^n \bX_i^T \SSigma^{-1} \Eta_i  \right\|_* \leq \frac{2c_0 c_1 c_2  \cdot \kappa \sqrt{\mu_{\max}} }{\sqrt{n}}  \cdot \sqrt{ \tr\left(\SSigma^{-1}  \SSigma_* \SSigma^{-1}\right)}  \cdot w(\cB)
\eeq
Finally we complete the proof by letting $C = 2c_0 c_1 c_2$, $C_1 = c_1$, and $C_2 = c_3$. \qed

\subsection{Proof of Theorem \ref{the:gds_bound}}

\proof By Corollary \ref{cor:re}, we have the RE condition hold with $\alpha = \frac{\mu_{\min}}{2} \cdot \tr(\SSigma^{-1})$ for $\cA(\Th^*)$.
Combining Lemma \ref{lem:gds} and \ref{lem:gamma_n}, we get
\begin{align}
\|\hat{\Th} - \Th^*\|_2 \leq 2\Psi(\Th^*) \cdot \frac{\gamma_n}{\alpha} 
\leq C\kappa \sqrt{\frac{\mu_{\max}}{\mu_{\min}^2}}  \cdot \frac{\sqrt{ \tr\left(\SSigma^{-1}  \SSigma_* \SSigma^{-1}\right)}}{\tr\left(\SSigma^{-1}\right)} \cdot \frac{\Psi(\Th^*) \cdot w(\cB)}{\sqrt{n}}  ~,
\end{align}
and the probability is computed via union bound. \qed

\subsection{Proof of Lemma \ref{lem:re_special}}

\proof Let $\tilde{\x}^{j^T}_i$ denote the $j$-th row of $\tilde{\bX}_i$, which is identically distributed as $\tilde{\x}$. In order to use Lemma \ref{lem:mept}, we let $(\Omega, \mu)$ be the probability measure that $\tilde{\x}$ is defined on. Construct the set of points $\cA_{\LLambda}$ according to \eqref{eq:re_set} and the function set
\begin{align*}
\cH = \left\{ h_{\v} = \langle \v, \cdot \rangle \  | \ \v \in \cA_{\LLambda} \right\}
\end{align*}
Since $\cA_{\LLambda} \subseteq \s^{p-1}$ and $\tilde{\x}$ is isotropic, it is easy to verify that $\E[h^2_{\v}] = \E_{\tilde{\x} \sim \mu}[\langle \tilde{\x}, \v \rangle^2] = 1$, $\vertiii{h_{\v}}_{\psi_2} \leq \tilde{\kappa}$ for every $h_{\v} \in \cH$, and $\vertiii{h_{\v_1} - h_{\v_2}}_{\psi_2} \leq \tilde{\kappa} \|\v_1-\v_2\|_2$ for any $h_{\v_1}, h_{\v_2} \in \cH$. Further, if we let $\beta = \frac{1}{2}$ and $mn \geq 4 c_1 c_2 \tilde{\kappa}^4 w^2(\cA_{\LLambda}) \triangleq C_1 \tilde{\kappa}^4 w^2(\cA_{\LLambda})$, using \eqref{eq:gamma2_1}, \eqref{eq:gamma2_2} and \eqref{eq:gamma2_3}, we have
\begin{align*}
c_1 \tilde{\kappa} \gamma_2\left(\cH, \vertiii{\cdot}_{\psi_2}\right) \leq c_1 \tilde{\kappa} \gamma_2\left(\cA_{\LLambda}, \|\cdot\|_2 \right) \leq c_1 c_4 \tilde{\kappa}^2 w\left(\cA_{\LLambda} \right) \leq \beta \sqrt{mn}
\end{align*}
By Lemma \ref{lem:mept}, with probability at least $1 - \exp(-c_2 \beta^2 mn / \tilde{\kappa}^4) \triangleq 1 - \exp(-C_2 mn / \tilde{\kappa}^4)$,
\begin{gather*}
\sup_{h \in \cH_j} \left|\frac{1}{mn} \sum_{i=1}^n \sum_{j=1}^m h^2(\tilde{\x}^j_i) - \E[h^2] \right| =
\sup_{\v \in \cA_{\LLambda}} \left|\frac{1}{mn} \sum_{i=1}^n \v^T \tilde{\bX}_i^T \tilde{\bX}_i \v - 1 \right| \leq \frac{1}{2} \\
\Longrightarrow \quad  \frac{1}{n} \sum_{i=1}^n \v^T \tilde{\bX}_i^T \tilde{\bX}_i \v  \geq \frac{m}{2}, \quad \forall \ \v \in \cA_{\GGamma_j} \\
\Longrightarrow \quad  \frac{1}{n} \sum_{i=1}^n \v^T \tilde{\bX}_i^T \XXi^{\frac{1}{2}} \SSigma^{-1} \XXi^{\frac{1}{2}} \tilde{\bX}_i \v  \geq \frac{m}{2} \cdot \lambda_{\min}\left( \XXi^{\frac{1}{2}} \SSigma^{-1} \XXi^{\frac{1}{2}} \right), \quad \forall \ \v \in \cA_{\LLambda}    \\
\Longrightarrow \ \  \frac{1}{n} \sum_{i=1}^n \v^T \LLambda^{-\frac{1}{2}} \LLambda^{\frac{1}{2}} \tilde{\bX}_i^T \XXi^{\frac{1}{2}} \SSigma^{-1} \XXi^{\frac{1}{2}} \tilde{\bX}_i \LLambda^{\frac{1}{2}} \LLambda^{-\frac{1}{2}} \v  \geq \frac{m}{2} \cdot \lambda_{\min}\left( \XXi^{\frac{1}{2}} \SSigma^{-1} \XXi^{\frac{1}{2}} \right) \v^T \v , \ \ \forall \ \v \in \cA_{\LLambda}
\end{gather*}
Now we replace $\LLambda^{-\frac{1}{2}} \v$ by $\w$ and use the definition of $\cA_{\LLambda}$ to obtain
\begin{gather*}
\frac{1}{n} \sum_{i=1}^n \w^T \LLambda^{\frac{1}{2}} \tilde{\bX}_i^T \XXi^{\frac{1}{2}} \SSigma^{-1} \XXi^{\frac{1}{2}} \tilde{\bX}_i \LLambda^{\frac{1}{2}} \w  \geq \frac{m}{2} \cdot \lambda_{\min}\left( \XXi^{\frac{1}{2}} \SSigma^{-1} \XXi^{\frac{1}{2}} \right) \cdot \w^T \LLambda \w , \quad \forall \ \w \in \cone(\cA) \\
\Longrightarrow \quad  \frac{1}{n} \sum_{i=1}^n \w^T \bX_i^T  \SSigma^{-1} \bX_i \w  \geq \frac{m}{2} \cdot \lambda_{\min}\left( \XXi^{\frac{1}{2}} \SSigma^{-1} \XXi^{\frac{1}{2}} \right) \cdot \lambda_{\min}\left(\LLambda\right) , \quad \forall \ \w \in \cA \\
\Longrightarrow \quad \w^T \hat{\GGamma} \w \geq \frac{m}{2} \cdot \lambda_{\min}\left( \XXi^{\frac{1}{2}} \SSigma^{-1} \XXi^{\frac{1}{2}} \right) \cdot \lambda_{\min}\left(\LLambda\right) , \quad \forall \ \w \in \cA
\end{gather*}
Finally we need to bound the Gaussian width $w(\cA_{\LLambda})$. Note that the proof of Lemma \ref{lem:subgauss_exp} implies that $\|\XXi^{\frac{1}{2}} \u\|_2^2  \cdot \LLambda = \E[\bX^T \u \u^T \bX] = \GGamma_{\u}$ for any $\u \in \s^{p-1}$. Therefore it is not difficult to see that $\cA_{\LLambda} = \cA_{\GGamma_{\u}}$. Using Lemma \ref{lem:subgauss_exp} and \ref{lem:width_rel}, we have
\begin{align*}
w(\cA_{\LLambda}) = w(\cA_{\GGamma_{\u}}) \leq C \kappa_0 \sqrt{\frac{\mu_{\max}}{\mu_{\min}}} \cdot \left( w(\cA) + 3 \right) = C \kappa_0 \sqrt{\frac{\lambda_{\max}(\XXi)\lambda_{\max}(\LLambda)}{\lambda_{\min}(\XXi)\lambda_{\min}(\LLambda)}} \cdot \left( w(\cA) + 3 \right) ~,
\end{align*}
which completes the proof. \qed

\subsection{Proof of Corollary \ref{cor:single}}

\proof Setting $n = 1$ and $\SSigma = \SSigma_* = \bI$ for Lemma \ref{lem:gamma_n}, we have
\begin{align*}
\left\| \bX^T \SSigma^{-1} \Eta  \right\|_* = \left\| \bX^T \Eta  \right\|_* \leq c \tilde{\kappa}  \sqrt{m \cdot \mu_{\max}}  \cdot w(\cB) = c \tilde{\kappa}  \sqrt{m  \cdot \lambda_{\max}(\XXi) \lambda_{\max}(\LLambda)}  \cdot w(\cB)~,
\end{align*}
with probability $1 - \exp\left(-\frac{m}{2}\right) - C_2\exp(-\frac{C_1^2 w^2(\cB)}{4 \rho^2})$. By Lemma \ref{lem:re_special}, we have $\alpha = \frac{m \cdot \lambda_{\min}\left( \XXi \right)  \lambda_{\min}(\LLambda)}{2}$, with probability at least $1 - \exp(-C_3 m / \tilde{\kappa}^4)$. Therefore, it follows from Lemma \ref{lem:gds} that
\begin{align*}
\|\hat{\Th}_{\text{sg}} - \Th^*\|_2 \leq 2\Psi(\Th^*) \cdot \frac{\gamma}{\alpha} \leq  C \tilde{\kappa} \cdot \sqrt{ \frac{\lambda_{\max}(\XXi) \lambda_{\max}(\LLambda)}{\lambda^2_{\min}(\XXi) \lambda^2_{\min}(\LLambda)}} \cdot \frac{\Psi(\Th^*) \cdot w(\cB)}{\sqrt{m}}
\end{align*}
which completes the proof. \qed

\section{Proofs for Section \ref{sec:cov_analysis}}
\subsection{Proof of Theorem \ref{the:est_cov}}

\proof By introducing the true parameter $\Th^*$, $\hat{\SSigma}$ can be rewritten as
\begin{align*}
\hat{\SSigma} &= \frac{1}{n} \sum_{i=1}^n \left(\Eta_i + \bX_i(\Th^* - \Th)\right) \left(\Eta_i + \bX_i(\Th^* - \Th)\right)^T
\end{align*}
And note that
\begin{align*}
\SSigma_{\Th} \triangleq \E[\hat\SSigma] = \SSigma_* + \DDelta_{\Th}, \ \ \text{where} \ \ \DDelta_{\Th} = \E\left[\bX(\Th^* - \Th)(\Th^* - \Th)^T \bX^T\right].
\end{align*}
The $\psi_2$-norm of $\SSigma_*^{-\frac{1}{2}} \left(\Eta + \bX(\Th^* - \Th)\right)$ satisfies
\begin{align*}
\vertiii{\SSigma_*^{-\frac{1}{2}} \left( \Eta + \bX(\Th^* - \Th) \right)}_{\psi_2} &\leq \vertiii{\SSigma_*^{-\frac{1}{2}} \Eta}_{\psi_2} + \vertiii{\SSigma_*^{-\frac{1}{2}} \bX(\Th^* - \Th) }_{\psi_2} \\
&=  \vertiii{\tilde{\Eta}}_{\psi_2} + \sup_{\u \in \s^{m-1}}\vertiii{(\Th^* - \Th)^T  \GGamma_{*\u}^{\frac{1}{2}}  \GGamma_{*\u}^{-\frac{1}{2}} \bX^T  \SSigma_*^{-\frac{1}{2}} \u}_{\psi_2} \\
&\leq \kappa_0 +  \sup_{\substack{\v \in \s^{p-1} \\ \u \in \s^{m-1}}} \left\|  \GGamma_{*\u}^{\frac{1}{2}} (\Th^* - \Th) \right\|_2 \cdot \vertiii{\v^T \GGamma_{*\u}^{-\frac{1}{2}} \bX^T  \SSigma_*^{-\frac{1}{2}} \u}_{\psi_2} \\
&\leq \kappa_0 +  \kappa \sup_{\substack{ \u \in \s^{m-1}}} \left\|  \GGamma_{*\u}^{\frac{1}{2}} \right\|_2 \left\|\Th^* - \Th \right\|_2  \\
&\leq \kappa_0 + \kappa \sqrt{\frac{\mu_{\max}}{\lambda_{\min}\left( \SSigma_* \right)}} \left\|\Th^* - \Th\right\|_2
\end{align*}
where $\GGamma_{*\u} = \E [\bX^T \SSigma_*^{-\frac{1}{2}} \u \u^T \SSigma_*^{-\frac{1}{2}} \bX]$, and $\|\GGamma_{*\u}\|_2^2 \leq  \mu_{\max}\|\SSigma_*^{-\frac{1}{2}} \u \|_2^2 \leq \frac{\mu_{\max}}{\lambda_{\min}(\SSigma_*)}$ by the definition of sub-Gaussian $\bX$. $\kappa_0$ is the $\psi_2$-norm of standard Gaussian random vector.
By Theorem 5.39 and Remark 5.40 in \cite{vers12}, if $n \geq C_0^4 m \left(\kappa_0 + \kappa \sqrt{\frac{\mu_{\max}}{{\lambda_{\min}\left( \SSigma_* \right)}}} \left\|\Th^* - \Th\right\|_2 \right)^4 $, with probability at least $1 - 2\exp(- C_1 m)$, we have
\begin{align}
\label{eq:cov_1}
\begin{split}
\left\|\SSigma_*^{-\frac{1}{2}} \left(\hat{\SSigma} - \SSigma_{\Th} \right) \SSigma_*^{-\frac{1}{2}} \right\|_2 &\leq C_0^2 \left(\kappa_0 + \kappa \sqrt{\frac{\mu_{\max}}{{\lambda_{\min}\left( \SSigma_* \right)}}} \left\|\Th^* - \Th\right\|_2 \right)^2 \sqrt{\frac{m}{n}}
\end{split}
\end{align}
Hence we have
\begin{align}
\label{eq:cov1}
\begin{split}
\lambda_{\max}\left(  \SSigma_*^{-\frac{1}{2}} \hat{\SSigma} \SSigma_*^{-\frac{1}{2}} \right) &= \left\|  \SSigma_*^{-\frac{1}{2}} \hat{\SSigma} \SSigma_*^{-\frac{1}{2}}\right\|_2 \leq 1 + \left\|  \SSigma_*^{-\frac{1}{2}} \left( \hat{\SSigma} - \SSigma_{\Th} \right) \SSigma_*^{-\frac{1}{2}}\right\|_2 + \left\|  \SSigma_*^{-\frac{1}{2}} \DDelta_{\Th} \SSigma_*^{-\frac{1}{2}}\right\|_2 \\
&\leq 1 + C_0^2 \left(\kappa_0 + \kappa \sqrt{\frac{\mu_{\max}}{{\lambda_{\min}\left( \SSigma_* \right)}}} \left\|\Th^* - \Th\right\|_2 \right)^2 \sqrt{\frac{m}{n}} + \frac{\mu_{\max}}{{\lambda_{\min}\left( \SSigma_* \right)}} \left\|\Th^* - \Th\right\|_2^2 \\
&\overset{(a)}{\leq} 1 + 2 C_0^2 \kappa_0^2 \sqrt{\frac{m}{n}} + \frac{ 2 C_0^2 \kappa^2 \mu_{\max}}{{\lambda_{\min}\left( \SSigma_* \right)}} \left\|\Th^* - \Th\right\|_2^2 \sqrt{\frac{m}{n}} + \frac{\mu_{\max}}{{\lambda_{\min}\left( \SSigma_* \right)}} \left\|\Th^* - \Th\right\|_2^2 \\
&\leq 1 + 2C_0^2 \kappa_0^2 \sqrt{\frac{m}{n}} + \left(\frac{\mu_{\min} }{{\lambda_{\max}\left( \SSigma_* \right)}} + \frac{\mu_{\max} }{{\lambda_{\min}\left( \SSigma_* \right)}} \right)\left\|\Th^* - \Th\right\|_2^2 \\
&\leq 1 + C^2 \kappa_0^2 \sqrt{\frac{m}{n}} + \frac{2\mu_{\max} }{{\lambda_{\min}\left( \SSigma_* \right)}} \left\|\Th^* - \Th\right\|_2^2
\end{split}
\\
\label{eq:cov2}
\begin{split}
\lambda_{\min}\left(  \SSigma_*^{-\frac{1}{2}} \hat{\SSigma} \SSigma_*^{-\frac{1}{2}} \right) &\geq 1 + \lambda_{\min}\left(  \SSigma_*^{-\frac{1}{2}} \left(\hat{\SSigma} - \SSigma_{\Th} \right) \SSigma_*^{-\frac{1}{2}} \right) + \lambda_{\min}\left(  \SSigma_*^{-\frac{1}{2}} \DDelta_{\Th} \SSigma_*^{-\frac{1}{2}} \right) \\
&\geq 1 - \left\|  \SSigma_*^{-\frac{1}{2}} \left(\hat{\SSigma} - \SSigma_{\Th} \right) \SSigma_*^{-\frac{1}{2}} \right\|_2 +  \frac{\mu_{\min}}{{\lambda_{\max}\left( \SSigma_* \right)}} \left\|\Th^* - \Th\right\|_2^2
\\ &\geq 1 - C_0^2 \left(\kappa_0 + \kappa \sqrt{\frac{\mu_{\max}}{{\lambda_{\min}\left( \SSigma_* \right)}}} \left\|\Th^* - \Th\right\|_2 \right)^2 \sqrt{\frac{m}{n}} +  \frac{\mu_{\min}}{{\lambda_{\max}\left( \SSigma_* \right)}} \left\|\Th^* - \Th\right\|_2^2  \\
&\overset{(b)}{\geq} 1 - 2C_0^2 \kappa_0^2 \sqrt{\frac{m}{n}} - \frac{2 C_0^2 \kappa^2\mu_{\max}}{\lambda_{\min}\left( \SSigma_* \right)} \left\|\Th^* - \Th\right\|_2^2 \sqrt{\frac{m}{n}} +  \frac{\mu_{\min}}{{\lambda_{\max}\left( \SSigma_* \right)}} \left\|\Th^* - \Th\right\|_2^2 \\
&\geq 1 - C^2 \kappa_0^2 \sqrt{\frac{m}{n}}
\end{split}
\end{align}
where $C^2 = 2C_0^2$, and in both (a) and (b), we use the assumption $n \geq C^4 m\kappa^4 \left(\frac{\lambda_{\max}\left( \SSigma_* \right)\mu_{\max}}{\lambda_{\min}\left( \SSigma_* \right)\mu_{\min}} \right)^2 = 4C_0^4 m\kappa^4 \left(\frac{\lambda_{\max}\left( \SSigma_* \right)\mu_{\max}}{\lambda_{\min}\left( \SSigma_* \right)\mu_{\min}} \right)^2$.
This completes the proof. \qed

\section{Proofs for Section \ref{sec:ae_analysis}}
\subsection{Proof of Lemma \ref{lem:factor_rel}}

\proof 
Based on the definition of $\xi(\cdot)$, we have
\begin{align}
\begin{split}
\xi\left(\hat{\SSigma}\right) &= \frac{\sqrt{ \tr\left(\hat{\SSigma}^{-1}  \SSigma_* \hat{\SSigma}^{-1}\right)}}{\tr\left(\hat{\SSigma}^{-1}\right)} = \frac{1}{\sqrt{\tr\left(\SSigma_*^{-1}\right)}} \cdot  \sqrt{\frac{\tr\left(\SSigma_*^{-1}\right) \cdot \tr\left(\hat{\SSigma}^{-1}  \SSigma_* \hat{\SSigma}^{-1}\right)}{\tr^2\left(\hat{\SSigma}^{-1}\right)}} \\
&=  \xi\left(\SSigma_*\right) \cdot  \sqrt{\frac{\tr\left(\hat{\SSigma}^{\frac{1}{2}} \SSigma_*^{-1} \hat{\SSigma}^{\frac{1}{2}} \hat{\SSigma}^{-1} \right) \cdot \tr\left(\hat{\SSigma}^{-\frac{1}{2}} \SSigma_* \hat{\SSigma}^{-\frac{1}{2}} \hat{\SSigma}^{-1}\right)  }{\tr^2\left(\hat{\SSigma}^{-1}\right)}} \\
&\leq  \xi\left(\SSigma_*\right) \cdot  \sqrt{\frac{\lambda_{\max}\left( \hat{\SSigma}^{\frac{1}{2}} \SSigma_*^{-1} \hat{\SSigma}^{\frac{1}{2}} \right)\tr\left(\hat{\SSigma}^{-1}\right) \cdot \lambda_{\max}\left(\hat{\SSigma}^{-\frac{1}{2}}  \SSigma_*   \hat{\SSigma}^{-\frac{1}{2}} \right) \tr\left(\hat{\SSigma}^{-1}\right) }{\tr^2\left(\hat{\SSigma}^{-1}\right)}} \\
&= \xi\left(\SSigma_*\right) \cdot  \sqrt{\lambda_{\max}\left( \hat{\SSigma}^{\frac{1}{2}} \SSigma_*^{-1} \hat{\SSigma}^{\frac{1}{2}} \right) \lambda_{\max}\left(\hat{\SSigma}^{-\frac{1}{2}} \SSigma_* \hat{\SSigma}^{-\frac{1}{2}} \right)} = \xi\left(\SSigma_*\right) \cdot  \sqrt{\frac{\lambda_{\max}\left( \SSigma_*^{-\frac{1}{2}} \hat{\SSigma} \SSigma_*^{-\frac{1}{2}} \right)}{\lambda_{\min}\left( \SSigma_*^{-\frac{1}{2}} \hat{\SSigma} \SSigma_*^{-\frac{1}{2}} \right)}}
\end{split}
\end{align}
where the inequality follows from von Neumann's trace inequality.
Now we can bound $\xi(\hat{\SSigma})$ by invoking Theorem \ref{the:est_cov},
\begin{align}
\begin{split}
\xi\left(\hat{\SSigma}\right) &\leq \xi\left(\SSigma_*\right) \cdot  \sqrt{\frac{1 + C^2 \kappa_0^2 \sqrt{\frac{m}{n}} + \frac{2\mu_{\max} }{{\lambda_{\min}\left( \SSigma_* \right)}} \left\|\Th^* - \Th\right\|_2^2}{ 1 - C^2 \kappa_0^2 \sqrt{\frac{m}{n}}}} \\
&= \xi\left(\SSigma_*\right) \cdot  \sqrt{1 + \frac{ 2 C^2 \kappa_0^2   \sqrt{\frac{m}{n}} + \frac{2\mu_{\max} }{{\lambda_{\min}\left( \SSigma_* \right)}} \left\|\Th^* - \Th\right\|_2^2}{ 1 - C^2 \kappa_0^2 \sqrt{\frac{m}{n}}}} \\
&\leq \xi\left(\SSigma_*\right) \cdot \left( 1 + \frac{\sqrt{2}C \kappa_0\left(\frac{m}{n} \right)^{\frac{1}{4}} + \sqrt{\frac{2\mu_{\max}}{{\lambda_{\min}\left( \SSigma_* \right)}}} \left\|\Th^* - \Th\right\|_2}{\sqrt{1 - C^2 \kappa_0^2 \sqrt{\frac{m}{n}}}} \right) \\
&\leq \xi\left(\SSigma_*\right) \cdot \left( 1 + 2C \kappa_0\left(\frac{m}{n} \right)^{\frac{1}{4}} + 2\sqrt{\frac{\mu_{\max}}{{\lambda_{\min}\left( \SSigma_* \right)}}} \left\|\Th^* - \Th\right\|_2 \right) \\
\end{split}
\end{align}
where the last inequality follows from $n \geq 4 C^4 m \cdot \left(\kappa_0 + \kappa \sqrt{\frac{\mu_{\max}}{{\lambda_{\min}\left( \SSigma_* \right)}}} \left\|\Th^* - \Th\right\|_2 \right)^4 \geq 4 C^4 m \kappa_0^4$. \qed

\subsection{Proof of Theorem \ref{the:alt_est}}

\proof Since $n \geq C^4 m \kappa^4 \left(\frac{\lambda_{\max}\left( \SSigma_* \right)\mu_{\max}}{\lambda_{\min}\left( \SSigma_* \right)\mu_{\min}} \right)^2$ and $\hat{\SSigma}_0$ is initialized as $\hat{\SSigma}_0 = \bI_{m \times m}$, by applying Theorem \ref{the:gds_bound} to $\hat{\Th}_1$, we have
\begin{align*}
\left\|\hat{\Th}_1 - \Th^*\right\|_2 &\leq C_1 \kappa \sqrt{\frac{\mu_{\max}}{\mu_{\min}^2}} \cdot  \xi\left(\hat{\SSigma}_0\right) \cdot\frac{\Psi(\Th^*) \cdot w(\cB)}{\sqrt{m}} = C_1 \kappa \sqrt{\frac{\mu_{\max}}{\mu_{\min}^2}} \cdot \frac{\Psi(\Th^*) \cdot w(\cB)}{\sqrt{mn}} \\
&\leq C_1 \kappa \sqrt{\frac{\mu_{\max}}{\mu_{\min}^2}} \cdot \frac{\Psi(\Th^*) \cdot w(\cB)}{\sqrt{m}} \cdot \frac{\lambda_{\min}\left( \SSigma_* \right)\mu_{\min}}{C^2 \sqrt{m} \cdot\kappa^2  \lambda_{\max}\left( \SSigma_* \right)\mu_{\max}} \\
&= \frac{C_1}{C^2} \cdot \frac{ \lambda_{\min}\left( \SSigma_* \right)}{\kappa  \lambda_{\max}\left( \SSigma_* \right) \sqrt{\mu_{\max}}} \cdot \frac{\Psi(\Th^*) \cdot w(\cB)}{m}
\end{align*}
It follows that
\begin{gather*}
n \geq C^4 m \cdot 4\left(\kappa_0 + \frac{C_1}{C^2} \sqrt{\frac{\lambda_{\min}\left( \SSigma_* \right)}{\lambda^2_{\max}\left( \SSigma_* \right)}} \frac{\Psi(\Th^*) w(\cB)}{m}\right)^4 \quad \Longrightarrow \\
n \geq C^4 m \cdot  4\left(\kappa_0 + \kappa \sqrt{\frac{\mu_{\max}}{{\lambda_{\min}\left( \SSigma_* \right)}}} \left\|\Th^* - \hat{\Th}_1\right\|_2 \right)^4
\end{gather*}
By applying Lemma \ref{lem:factor_rel} and Theorem \ref{the:gds_bound} to the second iteration,
\begin{gather*}
\left\|\hat{\Th}_{2} - \Th^*\right\|_2 \leq e_{\text{orc}} \cdot \left( 1 + 2C \kappa_0\left(\frac{m}{n} \right)^{\frac{1}{4}} + 2\sqrt{\frac{\mu_{\max}}{\lambda_{\min}\left( \SSigma_* \right)}} \left\|\hat{\Th}_1 - \Th^* \right\|_2 \right) \quad \Longrightarrow \\
\left\|\hat{\Th}_{2} - \Th^*\right\|_2 - e_{\text{min}} \ \leq \ 2 e_{\text{orc}} \sqrt{\frac{\mu_{\max}}{\lambda_{\min}\left( \SSigma_* \right)}} \cdot \left( \left\|\hat{\Th}_{1} - \Th^*\right\|_2 - e_{\text{min}} \right) ~.
\end{gather*}
Since $n \geq C^4m \cdot \left( \frac{2 C_1  \kappa}{C^2} \cdot \frac{\mu_{\max}}{\mu_{\min}} \cdot \frac{\xi(\SSigma_*) \Psi(\Th^*) w(\cB)}{\sqrt{m\cdot \lambda_{\min}(\SSigma_*)}} \right)^2$, we have $2 e_{\text{orc}} \sqrt{\frac{\mu_{\max}}{\lambda_{\min}\left( \SSigma_* \right)}} \ \leq \ 1$, which indicates that $\left\|\hat{\Th}_{2} - \Th^*\right\|_2 \leq \left\|\hat{\Th}_{1} - \Th^*\right\|_2$. Therefore the condition in Lemma \ref{lem:factor_rel} on sample size $n$ also holds for $\hat{\Th}_2$ and so on. By repeatedly applying Lemma \ref{lem:factor_rel} and Theorem \ref{the:gds_bound}, we have the following inequality for every $t > 0$,
\begin{align}
\label{eq:err_1}
\left\|\hat{\Th}_{t+1} - \Th^*\right\|_2 - e_{\text{min}} \ \leq \ 2 e_{\text{orc}} \sqrt{\frac{\mu_{\max}}{\lambda_{\min}\left( \SSigma_* \right)}} \cdot \left( \left\|\hat{\Th}_{t} - \Th^*\right\|_2 - e_{\text{min}} \right)
\end{align}
By combining \eqref{eq:err_1} for every $t$, we obtain
\begin{gather*}
\left\|\hat{\Th}_{T} - \Th^*\right\|_2 - e_{\text{min}} \ \leq \ \left(2 e_{\text{orc}} \sqrt{\frac{\mu_{\max}}{\lambda_{\min}\left( \SSigma_* \right)}}\right)^{T-1} \cdot \left( \left\|\hat{\Th}_1 - \Th^*\right\|_2 - e_{\text{min}} \right)
\end{gather*}
which completes the proof. \qed

\vspace*{3mm}
{\bf Acknowledgements:} The research was supported by NSF grants IIS-1447566, IIS-1422557, CCF-1451986, CNS-1314560, IIS-0953274, IIS-1029711, and by NASA grant NNX12AQ39A.


\bibliographystyle{plain}
\bibliography{ref}

\end{document}